\documentclass{article}
\usepackage[utf8]{inputenc}
\usepackage{ifthen}
\usepackage{titlesec}
\def\longversion{0}

\usepackage{microtype}
\usepackage{graphicx}
\usepackage{booktabs} 

\ifthenelse{\longversion=0}{
	\usepackage[accepted]{icml2018}
}{
	\usepackage{natbib}
	\usepackage{fullpage}
}

\usepackage{breakurl}

\usepackage{amssymb}
\usepackage{amsmath}
\usepackage{hhline}
\usepackage{url}

\usepackage{rotating}
\usepackage{ifthen}
\usepackage{algorithm}
\usepackage{algorithmic}
\usepackage{epsfig}
\usepackage{multirow}
\usepackage{array}
\usepackage{color}
\usepackage{multicol}
\usepackage{graphicx}
\usepackage{caption}
\usepackage{subcaption}
\usepackage{float}
\usepackage{hhline}
\usepackage{calc}
\usepackage{enumitem}
\usepackage{todonotes}
\usepackage{bold-extra}
\usepackage{ntheorem}
\usepackage{footnote}
\makesavenoteenv{tabular}
\makesavenoteenv{table}
\usepackage{threeparttable}
\usepackage{adjustbox}
\usepackage[normalem]{ulem}
\usepackage{mathtools}
\usepackage{booktabs}

\DeclarePairedDelimiter{\ceil}{\lceil}{\rceil}

\newtheorem{theorem}{Theorem}[section]
\newtheorem{corollary}{Corollary}[theorem]
\newtheorem{lemma}[theorem]{Lemma}

\def\dbar{\overline{D}}
\def\QED{\fbox{}}
\def\qed{\fbox{}}

\def\E{{\mathbb E}}

\def\Z{\mathbb{Z}}

\def\e{\epsilon }
\def\chi{{\mathbf 1}}

\def\R{{\mathbb R}}
\def\N{{\mathbb N}}

\newcommand{\TODO}[1]{\textcolor{red}{[todo:#1]}}

\ifthenelse{\longversion=0}{
\icmltitlerunning{Exact Random Forest with Billions of Examples}
}{}

\begin{document}

\ifthenelse{\longversion=1}{
{\center
\Large
{Exact Distributed Training: Random Forest with Billions of Examples\\ Supplementary material\\}
}}{

\twocolumn[
\icmltitle{Exact Distributed Training: Random Forest with Billions of Examples}





\begin{center}
\textbf{Mathieu Guillame-Bert and Olivier Teytaud}
\vspace{2mm}

Google, Zurich, Switzerland
\vspace{2mm}

\texttt{\{gbm,oteytaud\}@google.com}
\end{center}

\icmlkeywords{Machine Learning, Random Forest, Big Data, Distributed Algorithms}

\vskip 0.3in
]


}
\begin{abstract}
We introduce an exact distributed algorithm to train Random Forest models as
well as  other decision forest models without relying on approximating best split
search. We explain the proposed algorithm and compare it to related approaches for
various complexity measures (time, ram, disk, and network complexity analysis).
We report its running performances on artificial and real-world datasets of up to 18
billions examples. This figure is several orders of magnitude larger than datasets
tackled in the existing literature. Finally, we empirically show that Random
Forest benefits from being trained on more data, even in the case of already
gigantic datasets.
Given a dataset with 17.3B examples with 82 features (3 numerical, other categorical with high arity), our implementation trains a tree in 22h.
\end{abstract}

\sloppy
\setcounter{tocdepth}{5}
\ifthenelse{\longversion=1}{

\tableofcontents
}{}

\section{Introduction}
\ifthenelse{\longversion=0}{}{
Classification and regression consist in predicting respectively the class or the numerical label of an observation using a collection of training labelled records.
Decision Tree (DT) learning algorithms are a widely studied family of methods both for classification and regression~\cite{id3,c45}.
DTs have a great expression power (DT are universal approximators), they are fast to build (see complexity analysis in Section~\ref{sec:complexity}), and they are highly interpretable.
However, controlling DT overfitting is non-trivial.

DT bagging~\cite{randomforest, extremelyrandomizedtrees}, DT gradient-boosting~\cite{greedyboosting, stochasticgradientboosting} and DT boosting~\cite{adaboost} are three successful solutions aimed to tackle the DT overfitting problem. These methods (we refer to them as Decision Forest (DF) methods) consist in training collections of DTs. DF methods are state of the art for many classification and regression problems: as mentioned in \cite{yingdongbeatingkaggle, quorarf, tonghexgboost, tonghexgboost2}, the majority of winning methods in Kaggle contests~\cite{kaggle} are based on variants of Random Forests or Gradient Boosted Trees.

One notable domain where DF methods are generally not state of the art are computer vision problems, where Neural Networks have generally significantly higher performances. Exceptions include some use of Random Forest~\cite{ivan}, as well as the combining of DF with linear regression, leading to a deep and wide method as in \cite{fbads}.

As well as DT learning algorithms, generic DF methods~\cite{randomforest, extremelyrandomizedtrees, adaboost}
require a random memory access to the dataset during training. These methods are also non directly computationally distributable--the cost of network communication exceeding the gain of distribution.
These two constraints restrict the usage of DF methods to datasets fitting in the main memory of a single computer.
}
Two families of approaches have been studied and sometimes combined to tackle the
\ifthenelse{\longversion=1}{problem of training DTs and DFs on large datasets:}
{problem of training Decision Trees (DT) and Decision Forests (DF) on large datasets:}
(i) Approximating the building of the tree by using a subset of the dataset and/or
approximating the computation of the optimal splits with a cheaper or more easily
distributable computation, and (ii) using different but exact algorithms (building
the same models) that allow distributing the dataset and the computation.
\ifthenelse{\longversion=0}{
Various works\cite{dividedata, sliq} have shown that (i) typically leads to bigger forests and
lower precision.}{}
We focus on the latter family of approaches: we propose a distributed method which is exactly equivalent to the original DT algorithm. We compare our work to the two existing methods that fall in this category: Sprint~\cite{sprint} and distributed versions  of Sliq~\cite{sliq}.
\ifthenelse{\longversion=1}
{

	}{}
Our proposed method, inspired by Sliq, aims to reach:
(1) Removal of the random access memory requirement.
(2) Distributed training (distribution even of a single tree).
(3) Distribution of the training dataset (i.e. no worker requires access to the entire dataset).
(4) Minimal number of passes in terms of reading/writing on disk and network communication.
(5) Distributed computing of \emph{feature importance}.
\ifthenelse{\longversion=1}
{

	}{}
While this paper mainly focuses on Random Forests, the proposed algorithm can be applied to other DF models, notably Gradient Boosted Trees~\cite{sgbdt}.
\ifthenelse{\longversion=1}
{

	}{}
Our contributions in this work are as follows:
(1) A distributed and exact implementation of Random Forest able to train on datasets larger than in any such past work.
(2) A theoretical and numerical complexity comparison (CPU, RAM, network, disk access, disk reading, disk writing) of Sliq, Sprint, RF and our distributed version of Random Forest.

\ifthenelse{\longversion=0}{}{
The paper is structured as follows:
Section~\ref{sec:state_of_the_art} discusses related works.
Section~\ref{sec:drf} presents our Distributed Random Forest algorithm.
Section~\ref{sec:complexity} presents the theoretical complexity analysis of DRF and of related methods.
Sections~\ref{sec:experimentation} and \ref{sec:experimentation2} present experimental results.
We conclude the paper in Section~\ref{sec:conclusion}.
}

\ifthenelse{\longversion=0}{}{
\section{State of the Art}
\label{sec:state_of_the_art}

\subsection{Sequential Learning of Decision Trees}

In the sequential setting, a DT is build by recursively partitioning the training dataset according to the condition (also called \emph{split}) with the
highest immediate separative value for the labels. The tree is grown until the label of each leaf is pure,
or until a stopping criterion is met (e.g. maximum depth or minimum number of records in a leaf). Alg.~\ref{algo:decision_tree} shows the generic DT learning algorithm. Following the building process, the DT is then pruned to minimize the error estimated on an hold-out dataset.

\begin{algorithm}
\scriptsize
\textbf{BuildDecisionTree}($D$ : the training set)
\begin{algorithmic}
\STATE{Create a node $n$}
\STATE{\textbf{BuildNode}($D$, $n$)}
\RETURN $n$
\end{algorithmic}
\textbf{BuildNode}($D$ : a set, $n$ : a node)
\begin{algorithmic}
\STATE{\textbf{If} $D$ is pure \textbf{then} \textbf{return} }
\STATE{Find $c$ the condition with the highest separative value.}
\STATE{Label $n$ with $c$}
\STATE{Create $n'$ and $n''$ the two children of $n$.}
\STATE{Partition $D$ into $D'$ and $D''$ according to $c$.}
\STATE{\textbf{BuildNode}($D'$, $n'$)}
\STATE{\textbf{BuildNode}($D''$, $n''$)}
\end{algorithmic}

\caption{\label{algo:decision_tree} Generic decision tree learning algorithm. A condition is a split of the domain into 2 or more subdomains, often based on a single variable.}
\end{algorithm}

Different versions of DT~(CART~\cite{cart}, ID3~\cite{id3}, C4.5~\cite{c45}) algorithms differ on the type of supported conditions, the definition of the \textit{separative value} to maximize (i.e. splitting criterion), and the method (during or after training) to limit overfitting.

CART~\cite{cart} (Classification And Regression Trees) is a DT learning algorithm that builds binary trees such that each condition relies on a single attribute.
In CART, finding the optimal split for a numerical attribute is done by sorting the attribute values, and then by evaluating the splitting score of the mean of each two consecutive sorted values.
While this methods does not evaluate all possible threshold values (the support for such thresholds is infinite), it is guaranteed to have the highest splitting score.
We refer to this method as the \emph{exact numerical split} criterion.
In the case of categorical attributes, the optimal split can be found by a greedy selection based on the ranking of the possible values of the attribute. This method is optimal for binary classification but approximate for multiclass classification.
CART is used inside the Random Forest algorithm~\cite{randomforest}.

\subsection{Decision Forests}
Decision Forests is a generic term for models made of a collection of decision trees. Decision Forest algorithms generally differ on the training dataset used to feed a core DT algorithm, the type of noise injected when training DTs, the relations between the trees, and the constraints on the tree structure.

A Random Forest~\cite{randomforest} (RF) is a Decision Forest model defined as an ensemble of equally weighted DTs trained fully (i.e. without limit on the maximum depth) on independently bagged sets of examples (i.e. sampling with replacement; and generally with the same number of samples as the original dataset). More precisely, each tree is trained with CART~\cite{cart}. For each node, only a subset of candidate attributes is evaluated: for classification, the number of candidate attributes is often set to $\lceil \sqrt m\rceil$ with $m$ the total number of available attributes. The ensemble of trees in the RF compensates for the overfitting of each individual tree. Since each tree is trained independently, RF can be easily parallelized. However, training a RF requires for the entire dataset to be accessible with random access memory, hence the need for a distributed version working with a dataset distributed over many machines.

Gradient Boosted Trees~\cite{sgbdt} (GBT) is a Decision Forest model defined as a weighted list of decision trees where each tree is trained to predict the residual error of the previous trees. Once a tree is trained, its weight is determined by the resolution of a minimization problem; and then multiplied by a fixed constant called the shrinkage parameter (generally close to 0.1).
Unlike the trees of a RF, the trees in GBT are shallow (the maximum depth is generally set between 3 and 8). Because each tree in a GBT depends on its predecessors, the training computation cannot be parallelized as easily as RF.

Extremely Randomized Trees~\cite{extremelyrandomizedtrees} (ERT) is a popular DF algorithm that differs from RF on two core aspects. Trees learned by both algorithms share the same semantic but they are trained with a different type of noise: (1) RF trains each tree on a random bag (sampling with replacement) of the training examples. ERT uses all the available training examples for all the trees. (2) In RF, the splits for numerical attributes are selected from evaluating all possible meaningful threshold values
(i.e. the mean of each two contiguous sorted attribute values).
Instead, ERT considers a small (generally equal to the square root of the number of attributes, as for RF)
set of candidate attributes, and for each of them one single threshold sampled uniformly
between the minimum and the maximum values of this attribute in the current node.
This does not require to sort candidate numerical attributes, which make ERT nodes and trees faster to train than RF nodes and trees.
The noise used by ERT is stronger than the noise used by RF. This implies that each ERT split has a smaller separation power (on the training set) and a lower chance of overfitting (on the eval set) than a RF split.
Incidentally, ERT requires a larger number of trees to get the same performances of RF, but ERT also tends to converge to models with higher performances.

\subsection{Needs for Distributed Training}
The generic recursive DT learning algorithm described above requires for the entire
dataset to be available with random access memory. This prevents the algorithm
(including methods aimed at tackling large datasets such as \cite{megainduction} or \cite{window}) to
be used on large datasets that cannot fit in memory. And while each tree of a RF can be trained independently and
in parallel, GBT trees should be trained sequentially. These two factors prevent
usual decision forests algorithms to be executed on large datasets.

Two families of approaches have been studied for training decision trees on large datasets.
A first family of approaches relies on approximating the building of the tree by
using subsets of the dataset (at the tree level or at the node level) and/or by applying approximate,
less expensive and more easily distributable subroutines (notably to determine the optimal splits).
The second family of approaches relies on entirely different learning algorithms, natively distributable,
and with the guarantee of producing the exact same output as the recursive DT learning algorithm.

We discuss in details these two families of works in the next subsections.
Importantly, while we focus on random forest in its traditional variant, many hints
in the current work can be applied jointly with various methods for fastening random forest,
such as bag of little bootstraps\cite{bigdatab,rfbigdata},
extreme random forests\cite{extremelyrandomizedtrees},
or even with alternate forest methods\cite{sgbdt,tonghexgboost}.

%

\subsection{Large Scale Approximate Distributed Random Forest}
%
%
%
%
%
%
We here present various methods for approximate RF training,
besides approximate criteria such as extremely randomized trees mentioned above.

\subsubsection{Divide-and-conquer: approximate data-parallel methods.}
\cite{boat} uses a subset of data for constructing the first layers of the trees. Then, the
dataset is naturally split by these top layers and the lower layers can be tackled
separately. This approach is not easily applied in the context of forests, given
that even the first layers are different when subsets of features are considered.
E.g. \cite{desaidt,planet} (using Sparc or Mapreduce) do not have the same output as the original random forest or the original boosted decision
trees; most data-parallel methods\cite{dp1,dp2,dp3,dp4,planet} use approximate histograms and/or have a huge
computational cost as discussed in \cite{splitatt}. As discussed in some of these references and in
\cite{dividedata, sliq}, these approximate methods generate bigger forests and lose precision.
In addition of exactly matching the original random forest, without adding hyperparameters, we tackle datasets bigger, to the best of our knowledge,
than all published experiments - including training set sizes at which \cite{planet} explains that reaching exact splits is impossible.

\subsubsection{Divide and conquer: approximate feature-parallel methods.}
\cite{splitatt} splits datasets according to features, and restricts the flow of communication between the machines by a voting method for choosing the best attributes.
The resulting algorithm is not equivalent to full decision trees but huge datasets can be tackled in reasonable
time (11e6 examples and 1200 mostly numerical attributes in 5825s for one tree built on 8 machines; or 235e6
examples and 800 mostly discrete attributes in 5349s for one tree built on 32 machines).
The approach adds hyperparameters.

\subsubsection{Streaming methods.}
Some methods use streaming (e.g. \cite{vfdt,cvfdt}); usually they reach, asymptotically in the number of trees,
the same performance - but at the cost of bigger ensembles\cite{extremelyrandomizedtrees}.
\cite{bigrf1} reaches 1.2 millions examples in dimension 4K with an incremental method; this fits in memory.

\subsection{Exact Distributed Random Forests Based on Exact Distributed Decision Trees}
When working on datasets too big for being tackled on one single machine without a redhibitory number of I/O passes or memory consumption,
it makes sense to distribute the learning of each decision tree rather than just distributing the learning of the different trees.
Sprint\cite{sprint} and the distributed version of Sliq\cite{sliq} (called Sliq/R and Sliq/D) are the main approaches for computing exact decision trees in a distributed fashion.

\emph{Sprint}~\cite{sprint} first sorts each numerical column by values.
Each sorted column, called \emph{attribute list}, is stored on drive as a sorted list of tuples \textit{$<$sample index, attribute value, label value$>$}.
Sprint then learns a DT recursively, similarly as the generic DT algorithm, with the following modifications:
(1) Determining the optimal exact split of a numerical attribute only requires a single iteration over the
corresponding attribute list (i.e. no sorting is necessary).
(2) Once the optimal split is found, the attribute lists of both children nodes are
derived from the parent attribute lists and stored on drive. To do so, attribute lists
of the column used in the optional split are first computed.
At the same time, Sprints stores (in memory; in a hash table) the ids of the samples
that fail the optimal split condition i.e. the samples that will be sent to the
negative child.
The attribute lists of the two children for the remaining columns are then computed using this hash table.

\cite{sprint} (Sprint) is one of the most related works; it uses presorted attributes (presorting is also available
in standard libraries like SciKit\cite{scikit-learn}) in the context of random forest;
the original paper (1996) reports training a dataset of 1.6 millions examples using 20 processors.

As pointed out in \cite{scalparc}, the Achilles' heel of Sprint is the hashtable used concurrently. \cite{scalparc} presents an adaptation of Sprint with parallel hashtables learning one tree on 6.4 million records on 128 processors with 7 features in 77 seconds with no depth limit, on a high-quality infrastructure. In terms of large numbers of features, \cite{randomjungle} needs 0.53 hours for building 500 trees on 40 CPUs, with different number of features considered per node up to $19\times m/20$ where $m$ is the original number of features, $m=275153$, for 1006 samples.
\cite{cudt} presents a Cuda adaptation of Sprint.

Similarly to Sprint, \emph{Sliq}~\cite{sliq} first sorts each numerical column into an \emph{attribute list}.
For Sliq, an attribute list is a sorted list of \textit{$<$sample index, attribute value$>$} i.e. unlike Sprint the
attribute list does not contains the label values. Sliq then learns a tree ``depth level by depth level''
i.e. all the nodes at a given level are built at the same time, starting from the root node.
Sliq manages a dynamic mapping (in memory), called \emph{class list}, which maps each \textit{sample index}
to a \textit{node id} and \textit{label value}. At each stage, i.e. each depth level,
the attribute lists are scanned to determine the optimal split of each leaf (one single scan for all open nodes
of a given depth).
The class list is used to determine the node and the label corresponding to each sample.
Finally, the class list is updated according to the optimal splits, and the learning
continues to the next depth.

Sliq/R and Sliq/D~\cite{sprint} are two direct distributed version of Sliq.
In Sliq/R the attribute lists scanning is distributed attribute by attribute,
and the class list is replicated over all workers. In Sliq/D, samples are
distributed (shards), and the class list is distributed across workers.

\subsection{Comparison of Sliq and Sprint}
This section presents the main differences between Sliq~\cite{sliq} and Sprint~\cite{sprint}. See section~\ref{sec:complexity} for the in-depth complexity analysis of Sliq and Sprint.

In the generic DT algorithm, the computation bottleneck is the sorting of the numerical attributes that need to be repeated for each node. Both Sliq and Sprint avoid this cost by pre-sorting the numerical attributes i.e. the numerical attributes are sorted once.

For each node, Sprint computes and writes on disk the attribute list of each feature. In the case of a dense tree, since each entry of the attribute list contains three entries
\textit{$<$sample index, attribute value, label value$>$}, Sprint writes the equivalent of $3d$ times the training dataset on disk, with $d$ the depth of the tree. Instead, Sliq does not write on disk after the initial sorting of the numerical attributes.

In the case of attribute random sampling (e.g. RF), only a small subset of attributes are considered for the optimal split at each node. Sprint needs to write the attribute lists for both the selected and non selected attributes.

Sliq and Sliq/R need to maintain the entire class list in memory. The class list contains two entries (node index and label value) for each training sample. The class list can be stored as a dense array. Sprint needs to maintain a hash table in memory when splitting attribute lists. For the root node, and for a balanced tree, Sprint's hash table contains an entry for half of the training samples.
Sliq/D does not need to maintain the entire class list in memory.

Sprint distributes the training dataset sample by sample (i.e. row-wise) among its workers. When computing the optimal split, each worker computes the partial histogram (in the case of a categorical attribute) or the best split (in the case of a numerical attribute) of all the attributes. In the case of a categorical attribute, the partial histograms are aggregated and then used to compute the optimal split. In the case of a numerical attribute, each worker needs to know the label distribution of the samples with an attribute value lower and greater than the samples it owns. Therefore, for a given node, the communication between workers
depends on the number of workers and not on the number of training examples.

Like Sprint, Sliq/D distributes the class list and the training dataset sample by sample among its workers.
However, because each numerical attribute is sorted independently of others,
workers need to query each other's class lists' at the expense of a large network communication cost.

Sliq/R distributes the set of attribute lists among its workers - possibly one attribute list (corresponding to one feature) per worker.
After the optimal split has been computed, all the workers need to scan the attribute list of the attribute involved in the optimal split. If this is not possible, the worker responsible for this attribute list needs to send it to all the other workers.

At each depth level, Sliq scans the entire record of each candidate attribute.
When the depth of a tree increases, the \emph{sample density} (i.e. the ratio of samples owned by derivable leaves) decreases and the fraction of ignored records increases.
Instead, Sprint attribute lists only contain entries for the records in derivable leaves.

\def\justremovethis{
\subsection{Parallelization of boosted decision trees}
\TODO{ this section not ok}

\cite{sgbdt} parallelizes trees in the difficult context of stochastic gradient boosted decision trees, and provides
an exact implementation; they get positive results up to 20 machines, reducing the computation time from 70s to 9s with 1.2 millions examples
and 520 features, learning 10 trees until each of them has 20 leaf nodes. We deal with significantly bigger
training set sizes, and also provide exact solutions;
we focus on random forest, but our method is generic (we can even distribute the training
just for CART) and the application to a single tree can be used for gradient boosted trees as well.

\TODO{oteytaud: add the answers to my emails.}
}
\def\justremoved{
\TODO{Below is the older version}

\cite{sliq} takes care of saving up memory and presorts numerical attributes so that we do not need sorting at each level of the tree,
but still needs a linearly growing memory because a small amount of data must be kept in RAM for each training sample - but with a possibly very small constant.
Compared to Sliq, Sprint works in a fully distributed manner; no need for shared memory. It is usually considered as the main approach.
Nonetheless, Sliq works with reduced communications and less I/O sessions and can be implemented without shared memory.
The principle is to work per layer of the decision tree; Sliq splits nodes layer by layer. Sliq is the main inspiration for our work.

\cite{distxgboost} is very famous for its ability to work on large scale datasets; however, they switch to approximate methods
for their bigger dataset (1.7 billions records in dimension 67), explaining that their algorithm can not do
exact splits in such a case - we do so with 17 303 401 616 records and 82 attributes (79 categorical, 3 numerical, no sparsity).

\cite{sliq,sprint,scalparc} are the main related work; they use presorting, and they use attribute lists for reducing the memory cost.
Sprint became famous for the distributed case thanks to its reduced memory consumption by storing attribute lists in a distributed manner; it also reduces significantly the computational cost when the tree is highly unbalanced; but as shown in \cite{scalparc} the hashmap is an issue.
On the other hand,
Sliq does not need any reading/writing on disk except at the initialization and outperforms Sprint in the context of somehow shallow trees and/or when $m' << m$ i.e. in a random forest context. At the time of Sprint,
random forests were not as usual as nowadays; this significantly changes the comparison between both algorithms.
We use Sliq plus
\begin{itemize}
\item a rule for switching on the fly between Sliq and Sprint depending on the shape of the tree (section \ref{pmdt})
\item a method for sharing the bagging information for random forest within negligible network cost (using random seeds TODO)
\item a method with network cost one-bit-per-record method for updating class lists.
\end{itemize}

\subsubsection{\TODO{TODO; move the part about Sprint in the comparison section above} Boosted Decision Trees}
\cite{sgbdt} parallelizes trees in the difficult context of stochastic gradient boosted decision trees; they get positive results up to 20 machines, reducing the computation time from 70s to 9s with 1.2 millions examples and 520 features, learning 10 trees until each of them has 20 leaf nodes. \cite{sprint} (Sprint) is the most related work; it uses presorted attributes (presorting is available in standard libraries like SciKit\cite{scikit-learn}) in the context of random forest; the original paper (1996) uses 20 processors and reaches 1.6 millions examples. As pointed out in \cite{scalparc}, the Achilles' heel of Sprint is the hashtable used concurrently. \cite{scalparc} presents an adaptation of Sprint with parallel hashtables learning one tree on 6.4 million records on 128 processors with 7 features in 77 seconds with no depth limit, on a high-quality infrastructure. In terms of large numbers of features, \cite{randomjungle} needs 0.53 hours for building 500 trees on 40 CPUs, with different number of features considered per node up to $19\times m/20$ where $m$ is the original number of features, $m=275153$.
\cite{cudt} presents a Cuda adaptation of Sprint.


\section{\TODO{To merge with the related work and with the complexity analysis} Sliq and Sprint Algorithms}
\subsection{Description of algorithms}

%
}

\def\justremoved{

\subsection{\TODO{To move to the related work section} Pseudocodes}
Random Forest, Sprint and Sliq are presented in Alg. \ref{randomforest}, \ref{sprint} and \ref{mdt} respectively.

\TODO{Question: Do we need the RF algorithm? If so, we can use the (simpler) recursive DT learning algorithm listed in the related work section.
}
}

\begin{algorithm}
\scriptsize

\begin{algorithmic}[1]
\STATE{Inputs: $D_n\in \R^{n\times n_n}$ the input numerical attributes of the training dataset, $D_c\in \R^{n\times n_c}$ the categorical attributes of the training dataset, $D_\ell\in \N^n$ the categorical label of the training dataset. $D=D_n,D_c,D_\ell$.}
\FOR{Each tree $t$}
\STATE{$D'\leftarrow$ sampling of $n$ rows with replacement from $D$.}
\STATE{Open leaves $\leftarrow$ root associated to $D'$}
\WHILE{There exists an open leaf}
  \STATE{$\ell\leftarrow $ open leaf}
  \STATE{Randomly draw $m'$ attributes, where $m'=\sqrt{m}$}
  \STATE{$s\leftarrow$ best split over these attributes, or stop (depth or entropy or other stopping condition)}
  \STATE{Split on $s$: get two new leaves at most}
  \STATE{Close $\ell$}
\ENDWHILE
\ENDFOR
\end{algorithmic}
\caption{\label{randomforest} Random forest algorithm for classification. Small or homogeneous leaves are closed.}
\end{algorithm}
The complexity of learning, on one single core, one of the RF trees, with a straightforward implementation, is
$m'n_\ell\log n_\ell$ in node $\ell$; summing over the nodes yield an amortized $O(m' \dbar n\log(n))$ because each record is met on average $\dbar$ times, where $\dbar$ is the expected depth of leaves (weighted by the number of records per leaf). Summing over $T$ trees (for a sequential implementation) yields a complexity $O(Tm'\dbar n\log(n))$.
\begin{algorithm}
\scriptsize
\begin{algorithmic}
\STATE{Build attributes lists \& sort them\hfill $\lceil m/w \rceil n \log n$}
\STATE{Open leaves $\leftarrow$ root associated to $D'$}
\WHILE{There exists an open leaf}
  \STATE{$\ell\leftarrow $ open leaf}
  \STATE{Randomly draw $m'$ attributes, where $m'=\sqrt{m}$}
  \STATE{$s\leftarrow$ best split over these attributes\hfill $\lceil m'/w\rceil n D$}
  \STATE{Split on $s$: get two new leaves at most}
  \STATE{Send split info to all workers\hfill $\lceil m/w\rceil n\dbar$}
  \STATE{Each worker splits its attributes lists}
  \STATE{Close $\ell$}
\ENDWHILE
\end{algorithmic}
\caption{\label{sprint} Sprint for one tree. Complexities on the right hand side are summations
         over all loop iterations.}
\end{algorithm}
The overall complexity of sprint, assuming that the randomly drawn $m'$ attributes are approximately uniformly distributed (we will see in Section \ref{theory} that this assumption holds in many important cases), is $TK n\dbar + K n \log n$ for $T$ trees.
The memory requirement, for each worker, is linear in $n$ (storage of the hash map used for storing the split information), with a constant depending on the hashmap structure. We will see that Sliq, when distributed is competitive when $Z<<K$ or $\dbar \simeq D$.
\begin{algorithm}
\scriptsize
\begin{algorithmic}
\STATE{Build attributes lists \& sort them\hfill $\lceil m/w \rceil n \log n$}
\WHILE{There exists an open leaf, layer per layer}
  \FORALL{$\ell\leftarrow $ open leaf in current layer}
  \STATE{Randomly draw $m'$ attributes}
  \STATE{$s(\ell)\leftarrow$ best split over these attributes\hfill $\lceil zm'/w\rceil n D$}
  \ENDFOR
  \FORALL{$\ell\leftarrow $ open leaf in current layer}
  \STATE{Split $\ell$ on $s(\ell)$: get two new leaves at most for each $\ell$}
  \STATE{Send split info to all workers\hfill $D$ bitmaps of size $O(n)$ each}
  \STATE{Each worker updates its class lists}
  \ENDFOR
  \STATE{Close $\ell$}
\ENDWHILE
\end{algorithmic}
\caption{\label{mdt}Sliq decision tree. Complexities on the right hand side are summations over all loop iterations.}
\end{algorithm}
\begin{algorithm}
\scriptsize
\begin{algorithmic}
\STATE{Build attributes lists \& sort them\hfill $\lceil m/w \rceil n \log n$}
\WHILE{There exists an open leaf, layer per layer}
  \FORALL{$\ell\leftarrow $ open leaf in current layer}
  \STATE{Randomly draw $m'$ attributes}
  \STATE{$s(\ell)\leftarrow$ best split over these attributes\hfill $\lceil zm'/w\rceil n D$}
  \ENDFOR
  \FORALL{$\ell\leftarrow $ open leaf in current layer in ``no pruning'' mode}
  \STATE{Split $\ell$ on $s(\ell)$: get two new leaves at most for each $\ell$, in ``no pruning''}
  \STATE{Send split info to all workers}
  \STATE{Each worker updates its class lists}
  \ENDFOR
  \STATE{If Eq. \ref{pruning} holds then switch all nodes to the ``pruning'' mode.}
  \FORALL{$\ell\leftarrow $ open leaf in current layer in ``with pruning'' mode}
  \STATE{Split $\ell$ on $s(\ell)$: get two new leaves at most for each $\ell$, in ``with pruning'' mode.}
  \STATE{Send split info to all workers}
  \STATE{Each worker updates its attribute lists}
  \ENDFOR
  \STATE{Close $\ell$}
\ENDWHILE
\end{algorithmic}
\caption{\label{pmdt}One tree of DRF with pruning (DRFP). Complexities on the right hand side are summations over all loop iterations.}
\end{algorithm}
}

\section{Distributed Random Forest}
\label{sec:drf}
In this section, we describe the proposed Distributed Random Forest algorithm (DRF). The structure of this algorithm is different from the classical recursive Random Forest algorithm;
nonetheless, as well as Sliq~\cite{sliq} and Sprint~\cite{sprint}, the proposed algorithm is guaranteed to produce the same model as RF.
\ifthenelse{\longversion=1}
{

	}{}
DRF computation is distributed among computing units called ``workers'', and coordinated by a ``manager''. The manager and the workers communicate through a network.
\ifthenelse{\longversion=2}{In case of a large dataset, workers are instantiated on separate computers.}{} DRF is relatively insensitive to the latency of communication (\ifthenelse{\longversion=1}{see network complexity analysis in }{}Section~\ref{sec:complexity}).
\ifthenelse{\longversion=1}
{

	}{}
DRF also distributes the dataset between workers: each worker is assigned to a subset of columns \ifthenelse{\longversion=1}{(most often) or sometimes a subset of rows (for evaluators or if we add sharding)}{} of the dataset.
Each worker only needs to read their assigned part of the dataset sequentially, i.e. no random access and no writing are needed. Workers can be configured to load the dataset in memory, or to access the dataset on drive/through network access.
\ifthenelse{\longversion=1}
{

	}{}
Finally, each worker can host a certain number of threads \ifthenelse{\longversion=1}{. While workers communicate between each others through a network (with potentially high latency), we assume the threads of a given worker have access to a shared bank of memory. Most of the steps that compose DRF can be multithreaded.}{(details of multithreading in the supplementary material (SM)).}
\ifthenelse{\longversion=1}
{

	}{}
Several types of workers are responsible for different operations.
The \textbf{splitter} workers look for optimal candidate splits. Each splitter has access to a subset of dataset columns.
The \textbf{tree builder} workers hold the structure of one DT being trained (one DT per tree builder) and coordinate the work of the splitters. Tree builders do not have access to the dataset. One tree builder can control several splitters, and one splitter can be controlled by several tree builders.
\ifthenelse{\longversion=1}{The \textbf{OOB evaluator} workers evaluate continuously the out-of-bag (OOB) error of the entire forest trained so far. Each evaluator has access to a subset of the dataset rows.}{}
The \textbf{manager} manages the tree builders\ifthenelse{\longversion=1}{ and the evaluators}{}. The manager is responsible for the fully trained trees. The manager does not have access to the dataset.
Like Sliq, and unlike the generic DT learning algorithm, DRF builds DTs ``depth level by depth level'' i.e. all the nodes at a given depth are split together. The training of a single tree is distributed among the workers. Additionally, as trees of a Random Forest are independent, DRF trains all the trees in parallel. DRF can also be used to train co-dependent sets of trees (e.g. Boosted Decision Trees). In this case, while trees cannot be trained in parallel, the training of each individual tree is still distributed.
\ifthenelse{\longversion=1}{The following subsections provide the algorithm description and pseudocode.}{}
\subsection{Dataset Preparation}
\label{sec:ds_preparation}
\ifthenelse{\longversion=1}{
The first stage of the algorithm is about preparing the training set $D = \{(x_{i,j}, y_i); i=1,\cdots,n; j=1,\cdots,m\}$ where $n$ is the number of samples, and $m$ is the number of columns (also called attributes or features).
In this work, we consider each column to be either numerical or categorical.
First, a unique dense integer index is computed for each sample. If available, this index is simply the index $i$ of the sample in the dataset. Next, the dataset is re-ordered column-wise in increasing order of the sample indexes, and each  column is divided into $p$ shards: For each column, the shard $k$ contains the samples $i \in [t_k, t_{k+1}[$ with $t_{p+1}=n$. Finally, each numerical column is sorted by increasing attribute value. A sorted column is a list of tuples \textit{$<$attribute value, label value, sample index, (optionally) sample weight$>$}. Note: a sorted numerical column is similar to the \emph{attribute list} in Sprint~\cite{sprint}.}{
Consistently with existing works \cite{sliq,sprint}, we use presorting for numerical attributes. In the present work
we do not consider other categories of attributes than categorical or numerical.}
\ifthenelse{\longversion=1}{

}{}
The most expensive operation when preparing the dataset is the sorting of the numerical attributes. In case of large datasets, this operation is done using external sorting.
\ifthenelse{\longversion=1}{\subsection{Dataset Distribution}}{}
In this phase, the manager distributes the dataset among the splitters\ifthenelse{\longversion=1}{ and the evaluator workers}{}.
Each splitter is assigned with a subset of the dataset columns\ifthenelse{\longversion=1}{, and each evaluator is assigned with a subset of the dataset shards}{}.
In case several DTs are trained in parallel (e.g. RF), DRF benefits from having workers replicated i.e. several workers own the same part of the dataset and are able to perform the same computation.
\subsection{Seeding}\label{seeding}
\ifthenelse{\longversion=1}{
RF bags (sampling with replacement) samples used to build each tree: for each tree, each sample $i$ is selected $b_i$ times with $b_i$ sampled from the Binomial distribution
corresponding to $n$ trials with success probability $\frac{1}{n}$. Pre-computing and storing $b_i$ for each example is prohibitively expensive for large datasets. Instead, DRF computes $b_i$ on the fly using a fast pseudo random generator function: $b_i = \mbox{bag}(i,p)$ with $i$ the sample index and $p$ the tree index.
$\mbox{bag}(i,p)$ is a deterministic function.
DRF uses an implementation of $\mbox{bag}(i,p)$ as proposed in Alg. \ref{bagbag}.
This algorithm is a fixed number of steps of linear congruential generator that uses $i$ and $p$ as seeds.
This implementation is a low quality random generator, but it is fast and sufficient for the bagging task.
\begin{algorithm}
\scriptsize
\begin{algorithmic}
  \STATE{$a$, $b$ and $m$ are three fixed large prime numbers, and $n$ an integer (e.g. $n=3$).}
  \STATE{$k\mapsto\mbox{cdf}(k)$ is the cumulative distribution of the Binomial with $n$ trials and probability success $\frac{1}{n}$. $\mbox{cdf}(k)$ values are pre-computed for $k \in [0, K]$ (e.g. $K=10$).}
  \STATE{$c \leftarrow i$}
  \STATE{\textbf{for} $k\leftarrow 0, \cdots, n$ \textbf{do} $c \leftarrow (a c + b) \% m$}
  \STATE{$c \leftarrow c + p$}
  \STATE{\textbf{for} $k\leftarrow 0, \cdots, n$ \textbf{do} $c \leftarrow (a c + b) \% m$}
  \STATE{$v \leftarrow c / m$}
  \FORALL{$k\leftarrow 0, \cdots, K$}
  \STATE{\textbf{if} $v \leq \mbox{cdf}(k)$ \textbf{then} \textbf{returns} $k$}
  \ENDFOR
  \STATE{\textbf{returns} $K+1$}
\end{algorithmic}
\caption{\label{bagbag}Computation of $\mbox{bag}(i,p)$}
\end{algorithm}}{
RF ``bags'' samples (i.e. sampling with replacement, $n$ out of $n$ records) used to build each tree.
Instead of sending indices over the network, DRF uses a deterministic \ifthenelse{\longversion=1}{random-access }{}pseudo-random generator so that
all workers agree on the set of bagged examples without network communication.
}
With this method, all workers are aware of the selected samples,
without the cost of transmitting or storing this information.
\ifthenelse{\longversion=1}{
The random-access property removes the need for storing the samples in memory.
Similarly, Random Forest requires to select a random subset of
candidate attributes to evaluate at each node of each tree.
Following the same method, DRF uses the deterministic function
$\mbox{candidate}(j, h ,p)$, where $\mbox{candidate}(j, h ,p)$
specifies if the attribute $j$ is considered for the node $h$
of the tree $p$, and with $\mbox{candidate}(\N, \N ,\N)$ following
a binary distribution of success probability $\frac1{\sqrt{d}}$.
}{}
\subsection{Mapping Sample Indices to Node Indices}
At any point during training, each bagged sample is attached to a single leaf - initially the root node. When a leaf is derived into two children, each sample of this node is
re-assigned to one of its child nodes according to the result of the node condition (condition = chosen split).
As in Sliq~\cite{sliq}, DRF splitters and tree builders need to represent the mapping from
a sample index to a leaf node.
DRF monitors the number $\ell$ of active leaves (i.e. number of leaf nodes which can be further split).
Therefore, $\ceil{\log_2 \ell}$
bits of information are needed to index a leaf. If there is at least one non-active leaf,
$\ceil{\log_2 (\ell+1)}$ bits are needed to encore the case of a sample being in a closed leaf. Therefore,
this mapping requires $n \ceil{\log_2 (\ell+1)}$ bits of memory to store in which leaf each sample is.
\ifthenelse{\longversion=1}{

}{}
Depending on the size of the dataset, this mapping can either be stored entirely in memory, or the
mapping can be distributed among several chunks such that only one chunk is in memory at any time.
\ifthenelse{\longversion=1}{
The time complexity of DRF essentially increases linearly with the number of chunks for this mapping.

	}{}
Unlike Sliq~\cite{sliq}, DRF does not store the label values in memory.
\subsection{Finding the Best Split}
During training, each splitter is searching for the optimal split among the candidate attributes
it owns. The final optimal split is the best optimal split \ifthenelse{\longversion=1}{}{(e.g. for information gain or Gini index)} among all the splitters.
\ifthenelse{\longversion=1}{The optimal split is defined as the split with the highest split score.
DRF assumes the split score to be either the Information Gain or the Gini Index. }{}A split is defined as a column index $j$ and a condition over the values of this column. For numerical columns, the condition is of the form $x_{i,j} \leq \tau$ with $\tau \in \R$. For categorical columns, the condition is of the form $x_{i,j} \in C$ with $C \in 2^{S_j}$ and $S_j$ the support of column $j$.
In case of attribute sampling (e.g. RF), only a random subset of attributes are considered. 
We call \emph{supersplit} a set of splits mapped one-to-one with the open leaves at a given depth of a tree. The following subsections present how DRF computes the optimal splits
for all the nodes at a given depth, i.e. the \emph{optimal supersplit} at a given depth, in a single pass per feature.
Computing optimal splits on categorical attributes is easily parallelized, whereas computing optimal splits in the
case of numerical attributes needs presorting.
\ifthenelse{\longversion=0}{ Details are given in the SM for all cases, and
	Alg. \ref{algo:num_split} presents the algorithm for finding the optimal splits
  for a given numerical feature for all nodes of a given depth in one pass.
}{We now discuss these two cases.
\subsubsection{Categorical Attributes}
Estimating the best condition for a categorical attribute $j$ and in leaf $h$ requires
to compute the bi-variate histogram between the attribute values and the label values
for all the samples in $h$. \cite{cart} is then applied to find the optimal (in case of
binary labels) or approximate (in case of multiclass labels) split.

For a given categorical attribute $j$, given the mapping from the \textit{sample index}
to the \textit{open leaf index}, a splitter computes this bi-histogram for each of the open
leaves through a single sequential iteration on the records of the attribute $j$.
The listing is given in Alg.~\ref{algo:cat_split}. The iteration over the samples can
be trivially parallelized (multithreading over sharding).

\subsubsection{Numerical Attributes}
Estimating the exact best threshold $\tau$ for a numerical attribute requires
a sequential iteration over all the samples in increasing order of the attribute
values. Suppose $q(k, j, h)$ the sample index of the $k^{th}$ element sorted
according to the attribute $j$ in the node $h$ i.e.
$x_{q(0, j, h), j} \leq x_{q(1, j, h), j} \leq \cdots \leq x_{q(n_h-1, j, h), j}$.
During this iteration, the average of each two successive attribute values
$(x_{q(k, j, h), j}+x_{q(k+1, j, h), j})/2$ is a candidate values for $\tau$.
The score of each candidate is computed from the label values of the already
traversed samples, and the label values of the remaining samples.

For a given numerical attribute $j$, given the mapping from the \textit{sample index}
to \textit{open leaf index}, a splitter estimates the optimal threshold for each
of the open leaves through a single sequential iteration on the records ordered
according to the values of the attribute $j$. Since the records are already sorted
by attribute values (see section~\ref{sec:ds_preparation}), no
sorting is required for this step. The listing is given in Alg.~\ref{algo:num_split}.
\begin{algorithm}[t]
\scriptsize
\begin{algorithmic}
\STATE{$H_{h \in [1, \ell]}$ is an empty bi-histogram between the labels
     and the attribute $j$ for the leaf $h$}
\FORALL{$i$ in $1,\cdots,n$ \textit{\hfill// This loop can be parallelized}}
\STATE{$h \leftarrow \mbox{sample2node}(i)$}
\STATE{\textbf{if} $h$ is a closed node \textbf{then} \textbf{continue}}
\STATE{\textbf{if} $\mbox{candidate feature}(j, h ,p)$ is false \textbf{then} \textbf{continue}}
\STATE{$b \leftarrow \mbox{bag}(i,p)$ \textit{\hfill// Number of times $i$ is sampled in tree $p$}}
\STATE{\textbf{if} $b=0$ \textbf{then} \textbf{continue}}
\STATE{Add $(x_{i,j},y_i)$ weighted by $b$ to $H_h$}
\ENDFOR
\FORALL{open leaf $h$}
\STATE{Find best condition using bi-histogram $H_h$}
\ENDFOR
\end{algorithmic}
\caption{\label{algo:cat_split}Find the best supersplits for categorical attribute $j$ and tree $p$. Nodes are open when they are still subject to splitting - typically nodes are closed when they reach some purity level or when their cardinal is below some threshold.}
\end{algorithm}
}
\begin{algorithm}[t]
\scriptsize
\begin{algorithmic}
\STATE{$H_{h \in [1, \ell]}$ will be the histogram of the already traversed labels
      for the leaf $h$  (initially empty).}
\STATE{$v_{h \in [1, \ell]}$ is the last tested threshold (initially null) for the
       leaf $h$.}
\STATE{$q(j)$ is the list of records sorted according to the attribute $j$ i.e.
    $q(j)$ is a list of tuples $(a, y, i)$, sorted in increasing order of $a$,
    where $a$ is the numerical attribute value, $y$ is the label value,
    and $i$ is the sample index.}
\STATE{$t_{h \in [1, \ell]}$ will be the best threshold for leaf $h$  (initially null).}
\STATE{$s_{h \in [1, \ell]}$ will be the score of $t_h$  (initially 0).}
\FORALL{$(a, y, i)$ in $q(j)$}
\STATE{$h \leftarrow \mbox{sample2node}(i)$}
\STATE{\textbf{if} $h$ is a closed node \textbf{then} \textbf{continue}}
\STATE{\textbf{if} $\mbox{candidate feature}(j, h ,p)$ is false \textbf{then} \textbf{continue}}
\STATE{$b \leftarrow \mbox{bag}(i,p)$}
\STATE{\textbf{if} $b = 0$ \textbf{then} \textbf{continue}}
\STATE{$\tau \leftarrow (a+v_h)/2$}
\STATE{$s' \leftarrow$ the score of $\tau$ (computed using $H_h$)}
\IF{$s' > s_h$}
\STATE{$s_h \leftarrow s'$}
\STATE{$t_h \leftarrow \tau$}
\ENDIF
\STATE{Add label $y$ weighted by $b$ to $H_h$}
\STATE{$v_h \leftarrow a$}
\ENDFOR
\STATE{\textbf{return} $\{t_h\}$ and $\{s_h\}$}
\end{algorithmic}
\caption{\label{algo:num_split}Find the best supersplits for numerical attribute $j$ and tree $p$}
\end{algorithm}
\begin{algorithm}[t]
\scriptsize
\begin{algorithmic}[1]
\STATE Create a decision tree with only a root. Initially, the root is the only open leaf.
\STATE Initialize the mapping from \textit{sample index} to \textit{node index} so that all samples are assigned to the root. 
	\STATE Query the splitters for the optimal supersplit. Each splitter returns a partial optimal supersplit computed only from the columns it has access to (using Alg. \ref{algo:num_split} in the case of numerical splits). The (global) optimal supersplit is chosen by the tree builder by comparing the answers of the splitters.
\STATE Update the tree structure with the optimal supersplit.
\STATE Query the splitters for the evaluation of all the conditions in the best supersplit.
  Each splitter only evaluates the conditions it has found (if any).
  Each splitter sends the results to the tree builder as a dense bitmap. In total, all the splitters are sending one bit of information for each sample selected at least once in the bagging and still in an open leaf.
\STATE Compute the number of active leaves and update the mapping from \textit{sample index} to \textit{node index}.
\STATE Broadcast the evaluation of conditions to all the splitters so they can also update their \textit{sample index} to \textit{node index} mapping.
\STATE Close leaves with not enough records or no good conditions.
\STATE If at least one leaf remains open, go to step 3.
\STATE Send the DT to the manager.
\end{algorithmic}
\caption{\label{treebuilder}Tree builder algorithm for DRF.}
\end{algorithm}
\ifthenelse{\longversion=1}{
\subsection{Training a Decision Tree}
Each decision tree is built by the tree builder with Alg. \ref{treebuilder}.
\subsection{Training a Random Forest}
}{
\subsection{Training a Random Forest}
Each decision tree is built by the tree builder with Alg. \ref{treebuilder}.
}
To train a Random Forest, the \emph{manager} queries in parallel the \emph{tree builders}. This query contains the index of the requested tree (the tree index is used in the seeding, Section \ref{seeding}) as well as a list of splitters such that each column of the dataset is owned by at least one splitter. The answer by the tree builder is the decision tree.

\ifthenelse{\longversion=0}{}{
\subsection{Continuous Out-Of-Bag Evaluation}

The Out-Of-Bag (OOB) evaluation is the evaluation of a RF on the training dataset, such that each tree is only applied on samples excluded from their own bagging.
OOB evaluation allows to evaluate a RF without a validation dataset.
Computing continuously the OOB evaluation of a RF during training is an effective way to monitor the training and detect the convergence of the model.

During training, after the completion of each DT (or as specified by a walltime), the manager sends the new trees to a set of evaluators such that together, the set of evaluators covers the entire dataset (the dataset is distributed row-wise among the evaluators). Each evaluator then estimates the OOB evaluation of the RF on their samples. Evaluating $\mbox{bag}(i,p)$ on the fly, evaluators can detect if a particular sample $i$ was used to train a particular tree $p$. The partial OOB evaluation are then sent back to and aggregated by the manager.
The same method is used to compute the \emph{importance} of each feature.
}

\section{Complexity Analysis}\label{sec:complexity}
We present and compare the theoretical complexities (memory, parallel time, I/O and network) of generic DT, generic RF, DRF, Sprint, Sliq, Sliq/R and Sliq/D.
\ifthenelse{\longversion=1}{The main advantages of DRF over Sprint and Sliq/D-R are:


{\bf{A smaller memory consumption per worker}}; e.g., compared to Sprint, we reach, per worker,
   $num\ records \times (1 + \log_2 \max_{i}(num\ leaves\ at\ depth\ i))$ bits,
   instead of $num\ records \times sizeof(record\ index)$\ifthenelse{\longversion=1}{ with $sizeof(record\ index)$ equal to 64 bits for large datasets. Note: The memory consumption of DRF can be further reduced at the cost of an increase in time complexity.}{}

{\bf{A smaller amount and number of passes over data and of network communications.}} DRF's number of
passes over data and network communication is proportional to the depth of the tree; while it is proportional to the number of nodes for Sprint,
Sliq/D and Sliq/R.
The total number of exchanged bits is also smaller for DRF.
The network usage of Sliq/D is even greater since the node location of
each sample is only known by one worker, and since all the workers need access
to this information. DRF benefits from the communication
efficient synchronous sample bagging schema (Section \ref{seeding}).

{\bf{The absence of need for disk writing during training.}} DRF only writes on
disk during the initialization phase (unless the workers are configured to keep
the dataset in memory; in which case there are not disk writing at all).
In comparison, during training, Sprint writes on disk the equivalent of several
times the training dataset - for each tree in case of a forest.
}{}
All these algorithms operate differently, and benefit from different situations in term of time complexity:
\ifthenelse{\longversion=1}{

	}{}
Sprint prunes records in closed leaves: a tree with a large amount of records in shallow closed leaves is fast to train. However, Sprint scans and writes continuously both the candidate and non-candidate features i.e.
Sprint does not benefit from the small size of the set of candidate features.
%
%
\ifthenelse{\longversion=1}{

	}{}
Compared to Sprint, DRF benefits from records being in closed leaves differently:
records in closed leaves are not pruned, but since Sliq and DRF only scan candidate
features (i.e. features randomly chosen and not closed in earlier conditions),
a smaller number of records leads to a smaller number of candidate features.
Although our experiments focus on the classical case of features randomly drawn at each node,
we point out that Sliq and DRF benefit greatly
(by a factor proportional to the number of features) from limiting the number
of unique candidate features at a given depth. In particular, the trend (see Section \ref{maths}) consisting in using the same
set of features for all nodes at a given depth leads to a fast DRF
with a number of machines proportional
to the number of randomly drawn features instead of the total number of features.
\ifthenelse{\longversion=1}{

}{}
We also study the impact of equipping DRF with a mechanism to prune records similarly
to Sprint: when DRF detects that this pruning becomes beneficial,
the algorithm can prune the records in closed leaves.
This operation is not triggered during the experimentation on the large dataset reported in Section~\ref{sec:experimentation}.
\ifthenelse{\longversion=0}{}{
\subsection{Complexity in the sequential case with a dataset entirely stored in memory}
%
%
%
\ifthenelse{\longversion=1}{

	}{}
Sorting numerical attributes at every node is the bottleneck of training a DT with the generic sequential algorithm (see Alg.~\ref{algo:decision_tree}).
The worst case time complexity of this algorithm on $n$ training samples and $m$ numerical attributes is $O(m n^2 \log n)$~\cite{understandingrf}. The best case (balanced tree) and average case (training samples
have a 50\% chance to go in each branch) is
$O(m n \log^2 n)$~\cite{understandingrf}.
If we consider a maximum depth $\tilde d$ and a minimum number $p$ of records per node set by the user, the average tree depth becomes $\min(\tilde d, \log \frac{n}{p})$ and the time complexity of building a DT becomes
$O(m n \log n \min(\tilde d, \log \frac{n}{p}))$.
\ifthenelse{\longversion=1}{

	}{}
RF assumes a bagging (sampling with replacement) of the training examples and a sampling of $m' = \sqrt m$ attributes at each node. In this case, the average complexity analysis is
$O(\sqrt m \tilde{n} \log \tilde{n} \min(\tilde d, \log \frac{n}{p}))$ with $\tilde{n} \simeq 0.63 n$.
In the case of categorical only attributes, the average time complexity of RF is $O(\sqrt m \tilde{n} \min(\tilde d, \log \frac{n}{p}))$.
\ifthenelse{\longversion=1}{

	}{}
The generic sequential algorithm requires that the dataset is stored in memory. In the worst case, the memory complexity is $O(mn + n \min(d, \log \frac{n}{p}))$. In the average case, memory complexity is $O(mn)$.
}



\ifthenelse{\longversion=1}{
Compared to works mentioned above, the present paper considers distributed algorithms: we can deal
with data without having to store them in the memory of a single computer.
We report the time, parallel time, space and IO complexity analysis of the proposed algorithm DRF,
compared to Sliq and Sprint.}

\subsection{Distributed Forests: Complexity Overview}\label{over}
The present section compares the complexities of variants of distributed random forests; the formalization of our results is in Section \ref{theory}.
We mention Sprint\cite{sprint}, Sliq\cite{sliq}, but also Sliq/D and Sliq/R which were proposed
in \cite{sprint} as baselines for comparisons with Sprint. Sliq/D and Sliq/R are not designed
by the author of Sliq; DRF is another
solution (as opposed to Sliq/R and Sliq/D) for distributing Sliq.
We use workers (with distinct memories) for parallelizing the analysis of $m'$ features per node for a given depth,
these $m'$ features being randomly drawn (as usual in RF) out of the $m$ features of the dataset.
\label{notations}
In RF, the best split is chosen among splits of these $m'$ features.
There are variants of RFs for which these $m'$ features are the same for each node at
a given depth; this is in particular the case in the implementation Xgboost \cite{tonghexgboost} (which covers both
GBTs and RFs).
This has a big impact on the complexity in the distributed case, as discussed in Section~\ref{maths}.
Although our theoretical analysis and preliminary results suggest
a clear improvement with such a variant,
for the sake of an exact comparison with the most classical variant of random forest,
we will present experiments in the original case.
In the mathematical analysis, we use a variable $z$ counting the number of distinct subsets of $m'$ features drawn at a given depth; $z=1$ corresponds to the case in which
all nodes in a given depth level of the tree use the same set of $m'$ features, whereas $z$ is the total
number of open nodes of the current level in the case of independently drawn subsets of $m'$ features.
We call $n$ the number of samples in the dataset,
$m$ the number of attributes in the dataset,
$p$ the minimum number of records in a leaf,
$\tilde d$ the user-chosen maximum depth of a tree,
$D$ the effective depth of the tree i.e. depth of the deepest leaf ($D= \min(\tilde d, \log_2 \frac{n}{p})$ on average),
$\dbar$ is the average depth of the leaves (weighted by the number of training samples)
(by definition we have $\dbar \leq D\leq \tilde d$; $\tilde d$ is chosen by the user while $D$ and $\dbar$ are known after the tree is built;
equality means that the tree is perfectly balanced and the user limit is reached),
$w$ number of workers,
$T$ the number of trees to train,
$K = \lceil m/w \rceil$ is equal the maximum number of attributes owned by a worker when there is no redundancy
(we might have $\lceil dm/w\rceil$ features on a same worker in case of $d$-redundancies),
$n_\ell$ the number of training samples that reach node $\ell$,
$C$ the total number of nodes of a tree.

{\textsc{Data structures: comparing Sliq, Sprint and DRF.}}
Distributed versions of RF require to store (i) features and label values (i.e. the initial data, possibly after preprocessing such as distributing and presorting)
and (ii) class lists that maps training examples to open nodes.
For storing the data,
 Sprint and Sliq/D divide the data per rows (e.g. one shard per worker);
 Sliq/R and DRF divide the data per feature (i.e. a subset of features per worker).
For storing the class lists,
   Sliq/R and DRF duplicate the class list in each worker. To do so,
     DRF requires $\log_2(1+number\ of\ open\ nodes)$ bits per training example. In practice this figure is
     significantly smaller than using a full integer (i.e. 64 bits).
   Sliq/D and Sprint store a class list restricted to open nodes only.
      This saves up space/computation for later stages of the tree building,
      in particular when a large part of records
      are already in (closed) leaves. On the other hand,
      the class list contains the label (and possibly the weight) for
      each record, which is expensive, and we need many passes of writing, namely $\Omega(number\ of\ records)$ for
      each level of the tree. Sprint stores the class list as a distributed hashmap.

{\textsc{Communications in Sprint, in Sliq and in DRF.}}
    In Sliq/D, the class list is distributed over workers; this forces workers to
    query each others continuously, once per example.
    Sprint requests communications for updating the distributed hashmap, once per node.
    In DRF, for each supersplit (i.e. for each level of the tree),
    one bit of data is broadcasted for each training example in an open leaf.
    DRF and Sliq work with two passes per depth level, whereas Sprint, Sliq/D and Sliq/R work
    at the level of nodes.
In addition, for computing a random forest (compared to only a tree), we need bagging;
 instead of sending massive lists of record indices over the network, DRF just sends the seed
 of the randomized sampling (Section \ref{seeding}).

{\textsc{Computations and passes over the data in Sprint, in Sliq and in DRF.}}
For deciding the split,
Sprint, for categorical attributes, builds count tables ``attribute value $\times$ class $\to$  number of records''. For numerical attributes, incrementally compute the quality of splits (e.g. information gain), thanks to one histogram per class computed incrementally
     for each candidate threshold (i.e. each unique numerical attribute value) in order thanks to the presorted attributes - the histogram stores the number of individuals below the considered threshold, for each class.
     Sliq does the same, but for all open nodes of a given depth before broadcasting. For Sliq/D, based on shards, a large part of the cost is due to communications
     for combining histograms obtained on different shards.
 ~~For adapting the data to the splits,
 Sprint splits the attribute list for the chosen feature;
           collects split info in a hashmap; broadcasts this hashmap.
      Sliq: updates the mapping row id $\to$ node.
 DRF can be seen as an alternate solution for distributing Sliq (compared to Sliq/D and Sliq/R);
 as mentioned above, thanks to the same distribution of the data as in Sliq/R (per feature),
 we do not have to {\em{write}} anything regarding features (except during the initialization);
 features are simply read {\em{in one pass per level of the tree}} (and not per node of the tree!);
 and we broadcast one bit per record for updating the class lists, which are themselves stored
 with cost logarithmic in the number of open leaves.

\subsection{Complexity Analysis: Formalization}\label{theory}
\ifthenelse{\longversion=1}{
The present section provides formal elements for proving results in Table \ref{comptable}.
	\subsubsection{Overview of the Mathematical Lemmas and their Implications}}{}\label{maths}
{\textsc{$Z$: a critical quantity for the performance of DRF.}}
We distribute the $m$ features uniformly over the $w$ workers (i.e. at most $\lceil m/w\rceil$ features per worker) - though redundancies could be considered as detailed later. 
At each node, we randomly draw $m'$ features
for which optimal splits are computed;
the total number of drawn features is $m''\leq zm'$ with $z$ the number of independent
subsets of $m'$ features drawn in a given depth. Except in the USB case detailed below, $z$ is the
number of open nodes for the current depth.
The computational cost associated to computing splits for a node
is therefore, for a worker, proportional to the number of features which are
attributed to this worker.
We define $Z$ as the maximum number of features attributed to a given worker.
$Z$ depends on whether we apply USB;
on the number of features used in each node; on how many (and possibly which) workers have access to each feature (i.e. redundancies);
and on the number of workers. Several questions naturally arise in the analysis of $Z$.

{\textsc{Random forest with unique set of bagged features per depth (USB).}}
Importantly, the number $z$ of independently randomly drawn subsets of $m'$ features,
has a big impact on $Z$ and therefore on the overall complexity - we might consider a variant
of RF in which all nodes at a given depth consider a same subset of $m'$ features,
in which case we can set $z=1$ and $m''=m'$ independently of
the number of nodes. This feature, referred to as USB in the present document,
was already explored by \cite{tonghexgboost}.

{\textsc{Can we have $m''$ small at a given depth without having a narrow tree or USB ?}}
    Equivalently, we check if, without USB, features selected over the
    different open nodes in a single supersplit can have sufficient
		redundancies for reducing the total number of features to some $m''<<\min(zm',m)$.
		\ifthenelse{\longversion=1}{
		Lemma \ref{redundancies} shows that there is no hope - the number of selected features, up to a constant
		}{
		Lemmas in the SM show that there is no hope - the number of selected features, up to a constant
		}
		factor, verifies $\E m''=\Omega(\min( zm', m))$.

{\textsc{Correctness of the approximation ``nearly the same number of features are drawn on each worker''.}}
  If we have full redundancies (all features stored on all workers and optimal allocation of tasks to workers) or
  if $w=m$ then we clearly have a maximum number of features to be tackled on a given worker,
  for a given depth level, of the form $\E Z = O(\lceil m''/w\rceil)$ ($\E Z=1$ if $w=m$; and $\E Z=\lceil m''/w\rceil$
  with full redundancies ($d=w$) and optimal allocations);
  is there a risk, in the case $w<m$ and no redundancies, that the number of selected features is very unbalanced, so that one worker will
		take much more time than others ? Essentially,
		\ifthenelse{\longversion=1}{
			Lemma \ref{hypergeom} shows that this is not the case,}{
			Lemmas in the SM, based on
      VC-type inequalities for independent sampling\cite{laffertyconcentrationofmeasure}
      for the independent sampling case,
and
on VC-type inequalities for rejective sampling\cite{surveytrainingclemencon} for the non-independent case,
show that this is not the case}
		and the cost will remain $\E\Z =O(\lceil m''/w\rceil)$ if $m''$ increases ``faster'' than $w$;
    and remains $\E \Z=O(\log m'' / \log \log m'')$ with $w=m''$ even without redundancies\cite{Gonnet81}; details in SM.

{\textsc{In the case $w=m''$, a redundant storage of features improves the complexity in particular in the USB case.}}
  $w=m$ is for sure an ideal case (each worker deals with one and only one feature),
  but this might be impossible when $m$ is large; could we save up computational resources while
  preserving high performance when $m$ is large ? The answer, proved under the assumption of feature sampling with replacement,
  is yes.
     Assume $w=m''$, and let us store each feature on $d>1$ workers. Then,
		\ifthenelse{\longversion=1}{
		one of the cases in Lemma \ref{hypergeom} shows that instead of a complexity
		}{
		instead of a complexity
		}
    $\E Z$ of order $\log(m'')/\log \log m''$, we get $\log\log m'' / \log d$ (detailed proof in \cite{abku},
    more details in the SM).
    This, in conjunction with USB,
		leads to fast computation speed ($\E Z=O(1)$ if $\log(d)=\Omega(\log\log m)$)
    with $w=m'$ instead of $w=m$ - a significant improvement in
    the classical case $m'=\lceil \sqrt{m}\rceil$.

\ifthenelse{\longversion=0}{
}{
We also need proofs around external sort (for presorting numerical features);
for presorting, given that we work on distributed datasets which do not fit in
memory, we need so called external sort - we recall some
		known complexity bounds in Lemma \ref{externalsort}.

Using lemmas below, we get a complexity analysis as in
Table \ref{comptable}. In that table, please note that
we made assumptions as discussed above.

More precisely, $Z$ in Table \ref{comptable} is:
\begin{eqnarray}
Z&=&\lceil m'' / w \rceil \mbox{ if all features are available on all workers}\nonumber\\
    & & \mbox{and load balancing is optimized}\label{kp}\\
Z&=&O(\lceil m'' / w \rceil)\mbox{ if $m''$ increases faster than $w$ (see details below)}\nonumber\\
Z&=&O(\log(m'')/\log \log m'') \mbox{ if $m''=w$ and assuming}\nonumber\\ & &\mbox{feature sampling with replacement (FSWR).}\label{l3}\\
Z&=&O(\log(\log(m''))/\log(d) \mbox{ if $w=m''$ and FSWR and}\nonumber\\
& & \mbox{$d$-redundancies of features and}\nonumber\\
& & \mbox{greedy allocation.}\label{l4}
\end{eqnarray}
The first line corresponds to the case in which we distribute the load evenly over workers, using the assumption
that all features are available on all workers. Other lines show that we can reduce the number of workers while
preserving performance under some assumptions (the 2nd and 3rd line do not require any redundancy).
The case $w=m''$ is particularly interesting because it leads to all machines working full time (no wasted resource) and still
fast performance if $Z$ is moderate (e.g. $\log(d)=\Omega(\log\log m'')$ leads to $\E Z=O(1)$).
The greedy allocation refers to the following simple allocation process: for a given level, each time a feature is drawn, it is attributed
to the worker (among the $d$ workers which have access to it) with least load.


\subsubsection{Multivariate hypergeometric and multinomial distributions}
Let us consider $m''$ a sample size, which corresponds, for our applications,
  to the number of features which are drawn at a given depth.
If we bag features layer per layer (USB), then $m''=m'$; if we bag features node per node
then $m'' \leq zm'$. The main application is USB ($m''=m'$), and the case of a bounded $z$ (which can
be deduced from results below) - the case of no USB and $z$ large leads to $m''$ close to $m$ as discussed in Section \ref{mpplarge}.
We are interested in cases in which
\begin{equation}
\E Z \simeq \lceil m''/w\rceil \mbox{ i.e. } \E Z =\lceil m''/w\rceil+o(\lceil m''/w\rceil)\label{titi}
\end{equation}
or the weaker version
\begin{equation}
\E Z =O(\lceil m''/w\rceil )\label{titibis}
\end{equation}
or the even weaker version
\begin{equation}
\E Z =\tilde O(\lceil m''/w\rceil)\label{tititer}
\end{equation}
where $a=\tilde O(b)$ means $a=O((b+1)\times O(\log(b))$.
We will need the following properties of the multinomial and multivariate hypergeometric
distribution (the notations are chosen for matching the application to decision forests in the present document):

\begin{lemma}[Maxima of hypergeometric and multinomial variables]\label{hypergeom}
\ \\
Consider $w$ urns, with $K$ balls per urn. Let $m=w\times K$ be the total number of balls.
	Let us draw $m''$ balls, either with replacement ($r=1$) or with any rejective sampling
	(i.e. including a union of $z$ independent samplings without replacement of $m'$ features as in the sampling of features in a given depth level as in RF).
Let $z_i$ be the number of balls drawn in urn $i\in \{1,\dots,w\}$.
Consider $Z=\max_i z_i$ the maximum number of balls which have been drawn
(possibly with repetitions)
in any given urn.
We consider now $m$, $m''$ and $K$ depending on some integer
$n\in \{1,2,\dots,\infty\}$, so that $o(.)$ and $O(.)$ will make sense.

In the case $r=1$ (i.e. with replacement), $K$ has no impact, and
\begin{eqnarray}
\forall m'',w, \E Z &\leq& 2m''/w + 4m''^2w \exp\left(- \frac{3m''}{40w}\right).\label{tutu}\\
		    & & \mbox{(this implies Eq. \ref{titibis}}\nonumber\\
		    & & \mbox{if $wm''\exp\left(-3m''/(40w)\right)=o(1/w)$)}\nonumber\\
\forall m'',\mbox{ with }w=m'', \E Z &\simeq& \log m'' / \log\log m''\label{tototo}\\
   & &\mbox{ which implies that Eq. \ref{tititer} holds.}\nonumber
\end{eqnarray}
Still with replacement, if $m''=w$ and instead of randomly choosing an urn we choose $d$ urns
and we select the one in which the least number of balls have been drawn, then
\begin{equation}
\E Z = O(1) + (\log \log m'') / \log d.\label{tututu}
\end{equation}

In the case without replacement ($r=0$), we consider asymptotic results:
$w$, $K$ and $m'$ are functions of some $n\in \{1,2,\dots,\infty\}$.
If $w^2\exp(\log(w)\log(m+1)-m''/8w^2)=o(1)$, then
\begin{eqnarray}
\E Z &\leq& 2m''/w + o(m''/w)\mbox{(which implies Eq. \ref{titibis})}\label{eq2mw}
\end{eqnarray}
\end{lemma}
The first case with replacement (Eq. \ref{tutu}) and the case without replacement
(i.e. Eqs. \ref{eq2mw}) are new,
others are just reproduced here in a same framework
for convenience.

{\bf{Proof:}}
Let us consider Eq. \ref{tutu}.
Proving a convergence of $\E~ Z/m''$ to $1/w$ is easy in the case with replacement;
this is independent identical sampling so that the finiteness of the family of
indicator functions of urns (cardinal at most $w$) implies the finiteness of the shattering coefficients,
hence the frequency of each urn converges to its expectation.
More precisely, according to Theorem 7.86 in \cite{laffertyconcentrationofmeasure}:
$$P(Z/m''\geq 1/w + \e) \leq 2w \exp\left(-\frac{3m''\e^2}{24/w+4\e}\right)+2w\exp\left(-3m''\epsilon/40\right)$$
With $\e =1/w$, this yields
$$P(Z/m''\geq 2/w) \leq 4w\exp\left(-\frac{3m''}{40w}\right).$$

Using $\E Z/m'' \leq 2/w + P(Z/m''\geq 2/w)\max Z$ and $\max Z\leq m''$ ($\max Z$ refers to the maximum possible value for $Z$), this yields
$$\E Z/m'' \leq 2/w + 4m''w \exp\left(- \frac{3m''}{40w}\right),$$
hence Eq. \ref{tutu}.

An advantage of this approach is that a proposition in \cite{surveytrainingclemencon}
provides an adaptation of this argument to the case
without replacement. Let us recall this proposition below:

{\bf{Proposition 1 in \cite{surveytrainingclemencon}.}}

{\em{
Suppose that we sample $n$ points in a set of cardinal $N$ points with a rejective sampling scheme
with first order inclusion probabilities
$\pi_1,\dots,\pi_N$ and consider a class $C$ with finite VC-dimension $V$. Assume
$\kappa_N=(n/N) / \min_i \pi_i$. Then with probability at least $1-\delta$,
the supremum of the deviations between frequencies
and expectations in class $C$ is at most $U/3 + \sqrt{U}$, where
$U = 2\kappa_N [ \log(2/\delta) + V\log(N+1) ]  /  n$.
}}

{\bf{Proof:}} See \cite{surveytrainingclemencon}.\qed

We can apply the proposition above with $V=log(w)$, $\kappa_N=1$,
$n=m''$, $N=m$, $\pi_i=n/N$. We get that with probability $1-\delta$,
$Z/m'' \leq 1/w + U/3 + \sqrt{U}$ and where
$U = 2 [ \log(2/\delta) + \log(w)\log(m+1) ]  /  m''$,
\begin{equation}Z \leq m''/w + m''U/3 + m'\sqrt{U}.\label{yocool}\end{equation}

Let us choose $\delta=P(Z/m''\geq 2/w)$.
Eq. \ref{yocool} implies that
either $U \geq 1/4w^2$ or $2U/3 >= 1/w$.

Let us note $K=log(w)log(m+1)$.

{\bf{First case:}} if $U \geq 1/4w^2$,
$log(2/\delta) \geq m''/ (8w^2) - K$, hence
$2/\delta \geq exp(m''/ (8w^2) - K)$, hence
$\delta \leq 2exp( K -m''/ (8w^2) )$.

{\bf{Second case:}} If  $2U/3 >= 1/w$, hence
$\log(2/\delta) + K >= 3m'' / (4w)$, hence
$\delta \leq 2 exp ( K - 3m'' / (4w))$.

In both cases, $\delta \leq 2max\left( exp(K-3m''/4w), exp(K-m''/(8w^2))\right)$.
This implies (using $E Z \leq 2m''/w + \delta \max Z$ and $max Z \leq m''/w$),
$E Z \leq 2m''/w + 2max\left( exp(K-3m''/4w), exp(K-m''/(8w^2))\right) m''/w$.

This implies, if $w^2\exp(\log(w)\log(m+1)-m''/8w^2)=o(1)$, that
$E Z \leq 2m''/w + o(m''/w)$.


The proof that, with replacement, if $m''=w$, then $\E Z \simeq \log(m'') / \log(\log(m''))$ (Eq. \ref{tototo}) goes back to \cite{Gonnet81}.

The proof of the fourth case with replacement (Eq. \ref{tututu}) goes back to \cite{abku}.\qed

%


We now consider a setting in which we distribute $d$ replicas of each of the
$m$ features over the $w$ workers (we will see later on how to place the replicas).
Then, randomly drawn features are attributed to workers by minimizing the current load: more precisely,
for each feature,
there are $d$ replicas, placed on $d$ distinct workers; and for each randomly drawn feature, in turn, we choose the worker with minimum
load over these $d$ workers.

We have three random parts for evaluating $Z$ in this context:
\begin{itemize}
    \item the way replicas are distributed (initial attribution of replicas to workers);
    \item the way features are randomly drawn;
    \item the way we break ties when randomly choosing which of the replicas is used;
      this third random part has no impact on our results and we can actually decide
      an arbitrary order in case of ties.
\end{itemize}
Eq. \ref{tututu} looks like the solution for this case. However,
in a typical setting, the replicas are distributed once and only once, whereas Eq \ref{tututu} considers a real time
random choice of $d$ possible workers; therefore,
Eq. \ref{tututu} is not enough; we must check if there is a fixed attribution which provides a similar result.
This is proved as follows - we show that the probability of such a correct attribution is positive,
hence there exists such a correct attribution - and actually the result shows that we are likely to find
it by random choice.

\begin{corollary}[Replicas: redundant features]\label{coreg}
Assume that $m''=w$.

Initial attribution: randomly and independently place each of the $m''$ features,
each of them on $d$ of the $w$ workers,
each of these $m''$ random choices of $d$ workers being uniform over subsets of cardinal $d$.

Then, with probability at least $1-\delta$ over the initial attribution,
the expectation of the maximum number of features on a same worker when randomly
drawing $m''$ features and sequentially selecting one worker for each of them with
minimum load is $\leq \left(O(1) + \log \log(m'') / \log(d)\right)/\delta)$.
\end{corollary}

{\bf{Proof:}}
On average over the initial attribution and over the randomly drawn $m''$ features,
$\E_{initial\ attrib}\E_{features} Z = O(1) + \log \log(m'') / \log d$ thanks to Lemma \ref{hypergeom} (Eq. \ref{tututu}).

Then, by Markov's inequality, the probability over the initial attribution that
$\E_{features} Z > C (O(1)+\log \log(m'') / \log d)$ is at most $1/C$.
\QED

This means that, by selecting the right initial attribution, we switch from logarithmic in $m''$ to log-logarithmic in $m''$
if we have at least two replicas for each feature, in the case $w=m''$; and constant
if $w=m''$ and $\log d = \Omega(\log\log m'')$.

\subsubsection{External sort by multiway merge}
In the present document, we define a pass as an iteration over the entire dataset,
in the order the dataset is stored (on drive or in memory) i.e. no random or
multiple access to each piece of the dataset.
From e.g. \cite{extsort}, we know:

\begin{lemma}[External sort by multiway merge]\label{externalsort}
The total running time of multiway merge sort needs $O(n\log_{S-1}(\frac{n}{S-1}))$
I/O operations in $\log_{S-1}(\frac{n}{S})$ passes where $n$ is the number of disk
pages in the data and $S$ is the number of entries that can fit in memory.
\end{lemma}

\subsubsection{Repeated sampling: without USB, $m''$ is large}\label{mpplarge}
It is usual, in RF, to randomly draw $m'$ features, and to work only on these
$m'$ features - however, we can do this either with USB (i.e. randomly drawing $m'$ features once for all
nodes of a given depth) or bag at each node independently,
in which case $m''$ could be as large as $zm'$ - we here show that if $z$ is large (i.e.
without USB and no narrow tree), then $m''$ will be large - no good news with redundancies
between bagged sets of features, all features of all workers will be necessary.
More precisely, we show that the number of features used in a layer is, on average,
$\Omega(\min(zm',m))$ when we have $z$ open nodes and no USB.

\begin{lemma}[Redundancies over sets of features]\label{redundancies}
Consider $E$ a set of cardinal $m$.
Consider $E'_1,\dots,E'_z$ randomly chosen subsets of $E$, chosen independently
and uniformly among subsets of cardinal $m'$.
Consider $m''$ the cardinal of the union $E''$ of the $E'_i$.
Then $\E m'' \geq \min(zm'/2, m/2)$.
\end{lemma}

This lemma implies that the trivial bound $m''\leq \min(zm',m)$ can not be
improved except regarding the constants.
In particular, if, at each of $z$ nodes, we randomly sample $m'$ features among
$m$, we have to
select splits in $\Omega(\min(zm', m))$ features.

{\bf{Proof:}}
Let us note $p=m'/m$.
The probability that $e\in E$ is in $E''$ is $1-(1-p)^z$.
Therefore $1 - (\E m'')/m =(1-p)^z    \leq  1/(1+zp) $.
$ \E m'' \geq m-m/(1+zp) = (mzp)/(1+zp) \geq m'z / ( 1+zp) $.
Therefore $\E m'' \geq zm'/2$ if $z\leq m/m'$.
$\E m''$ increases as a function of $z$, therefore $\E m'' \geq m/2$ if $z\geq m/m'$.
\QED
}

\ifthenelse{\longversion=0}{
The key advantage of DRF is the moderate number of passes over the data. In case,
for some hardware, this would not matter {\em{and}} the pruning of Sprint might
perform particularly well because of many closed leaves early in the tree,
we can implement a rule for switching to Sprint's pruning
mode, and this rule detects the issue early enough for
preserving the complexity of Sprint in such a case (details in SM).
}{
\subsection{DRFP: the best of both world, DRF with pruning}\label{mdtp}
\label{sec:best_of_two_words}

Experimentally, and as shown in our complexity analysis (Table \ref{comptable}),
DRF performs well on large scale datasets because it has a reduced cost in terms of reading on disk
and limited memory cost. Let us see what happens if we consider only the computational cost, while neglecting
memory and disk reading/writing.

 For this context, DRF performs better or equivalently to sprint when
 \begin{equation}
 DZ=O(\dbar  K).\label{dkpdbk}
 \end{equation} This is essentially because:
\begin{itemize}
  \item $DZn$ is the total cost, for one worker of DRF, of computing for each required attribute and depth, the best split;
  \item $\dbar Kn$ is the total cost, for one worker of Sprint, of splitting each attribute.
\end{itemize}

Eq. \ref{dkpdbk} does not necessarily hold when the tree is not balanced and if we do not apply USB.
If we trust USB (which seems like a natural assumption, see Figure \ref{ubs}),
then Eq. \ref{dkpdbk} essentially concludes that DRF will outperform Sprint.
Let us also assume that the many writing passes of Sprint are not an issue;
after all, this could be hardware dependent.
Let us consider, then, what happens in such a context, i.e. we care only about computation (no IO passes)
and we want to stick to the original formulation of RF, i.e.
we randomly draw $m'$ features at each node. We show that we can implement in DRF
a rule for detecting the efficiency of Sprint, and switch to Sprint early enough for preserving its properties.

In concrete terms, we propose a version which automatically switches to pruned
attribute lists (i.e. attribute lists which are split explicitly, whereas
DRF just updates a class list). This is Alg. \ref{pmdt}. This can be applied with or without USB,
without assumption on the tree shape - it just ensures that we do not perform significantly worse than
Sprint even when pruning records is possible thanks to a highly unbalanced tree.

The principle is as follows.
Sprint's complexity is excellent when $\dbar=O(ZD/K)$, i.e. the tree is not shallow -
which makes sense for a pruning-based approach which only stores data corresponding to open leaves.
We therefore detect such situations by the following rule: switch to a pruning mode
(with attribute lists explicitly split)
for each tree such that the estimate of $\dbar$ is $\geq ZD/K$.
This rule is unlikely to be triggered
\begin{itemize}
\item if $\dbar$ and $D$ are of the same order (balanced tree);
\item or if $Z<<K$, which is usual in the random forests context if either
   the tree is narrow or USB.
\end{itemize}

More precisely, define ${\dbar}_i$ the expected depth of leaves of depth $\leq i$ and $\alpha_i$ the ratio of records which are still in open leaves at depth $i$.
Define $Z=\lceil z_im' /w\rceil$ where $z_i$ is the average number of open nodes for layers with depth $\leq i$.
We define DRFP, i.e. DRF with pruning, i.e. DRF switches to Sprint when we detect this to be beneficial,
as follows.
DRFP switches to pruning mode when
\begin{equation}
\underbrace{\alpha_i i + (1-\alpha_i)\dbar_i}_{\mbox{estimator of }\dbar} < Z_ii/K \label{pruning}.
\end{equation}

{\bf{First, let us see why DRFP performs as well as Sprint.}}
Consider $i$ such that we have not yet switched to Sprint.
The cost of Sprint (per worker, and besides the cost of the initial external sort), would we run it for the entire tree, is $$ Kn\dbar = \underbrace{K\alpha n i}_{A} + \underbrace{K\alpha n (\dbar_{i+}-i)}_{B}+ \underbrace{Kn\dbar_i(1-\alpha)}_C$$
where terms $A$ and $B$ refer to records still opened at depth $i$ and $C$ refers to records closed at depth $i$, and where $\alpha=\alpha_i$ and $\dbar_{i+}$ is the average depth of leaves at depth $>i$.

The cost of DRFP is
\begin{equation}
\underbrace{K\alpha n (\dbar_{i+}-i)}_{B} + \underbrace{Z_ini}_D,\label{mdtpcost}.
\end{equation}

We see that $B$ is already in the cost of Sprint. $D$ is $O(A+C)$ if $i$ is such that Eq. \ref{pruning} does not hold.
This means that if we could switch to pruning mode one step before Eq. \ref{pruning} holds, the cost of DRFP would be
$O(\mbox{cost of spring})$. We must check that the fact that we switch one layer too late is not a problem.
This is because the cost of one single layer of DRF is small compared to the cost of the first layer of Sprint;
so this also holds for the ``real'' DRFP which switches to pruning mode one step too late.

{\bf{Second: for a shallow tree, DRFP outperforms Sprint.}}
The discussion above has shown that DRFP performs as well as Sprint (up to constants); we now discuss cases in which it performs better.

In case of a perfectly balanced tree, $D=\dbar$, and Eq. \ref{pruning} never holds,
and (besides initial sorting) Sprint costs $KnD$ whereas DRF costs $ZnD$ - we save up a ratio $Z/K$.
With USB this is large as soon as Lemma \ref{hypergeom} holds; otherwise it is true
if and only if $z << m/m'$.

We left as further work the possibility to prune only parts of the nodes and/or switch back to non-pruning mode.
We refer to Table \ref{comptable} for a discussion including IO - actually the most critical
strength of DRF compared to Sprint.
}
{\begin{table*}[t]
{\center\scriptsize
\begin{tabular}{|p{2.3cm}||p{2.5cm}|p{2.2cm}|p{1.3cm}|p{3.3cm}|p{2.7cm}|}
\hline
Algorithm &
Max. memory (per worker) &
Computational cost (max per worker), i.e. parallel time complexity &
Writing on disk \& number of passes (per worker) &
Network \& number of passes &
Reading on disk \& number of passes (per worker) \\
\hline
\hline
Generic sequential recursive tree, all in memory &
 $m\times n\times [value]$ &
 $m'n\log(n)D$ &
 0 &
 0 &
	(m+1)n $[value]$ in $1$ pass \\
\hline
Sliq (on one machine) &
$n\times ([value] + [index\ of\ leaf])$ &
$m'' n D+ PS $ &
0 &
0 &
	$(m''+1)nD([value]+[record\ index])$ in $(m''+1)D$ passes \\
\hline
Sprint &
$n\times [record\ index]$
(nb: $[record\ index] \geq \log_2(n)\geq  \log_2(M)$)
&
$K n \dbar + PS $  &
PS + $C\times K$ passes of total size $Kn\dbar$ &
$n$ row indices for bagging + $\dbar n$ row indices in $C$ broadcasts; if we use bitmaps for saving up communication we will pay for sorting. &
$2Kn\dbar (2[value]+[record\ index])$ in $K\times C$  passes \\
\hline
Sliq/D   &
$\lceil n/w\rceil\times ([value]$ $+ [leaf\ index])$ &
$m'' \lceil n/w\ceil D+ PS $ and coordination &
PS  &
$n$ row indices for bagging and
coordination and $D$ broadcasts of $Dn$ bits. &
	$m''\lceil n/w\rceil D([value]+[record\ index])$ in $m''\times  C$ passes \\
\hline
Sliq/R & 
$n\times ([value]$ $+ [leaf\ index])$ &
$Z n D+ PS $ &
PS  &
$n$ row indices for bagging +
$Dn$ bits in $D$ allreduce. &
	$ZnD([value]+[record\ index])$ in $Z\times  C$ passes \\
\hline
DRF &
$n\times $ $(1+\log_2(M))$ &
	$(Z+1) n D+ PS $ &
PS  &
$Dn$ bits in $D$ allreduce. &
	$ZnD(2[value]+[record\ index])$ in $Z\times  D$ passes \\
\hline
DRF-USB, $w=m'$, $d=\log(m')$&
$n\times $ $(1+\log_2(M))$ &
$n D+ PS $ &
PS  &
$Dn$ bits in $D$ allreduce. &
$2Dn(2[value]+[record\ index])$ in $2D$ passes \\
\hline
\end{tabular}
}
\ifthenelse{\longversion=1}{
\caption{\label{comptable}Complexities of some discussed algorithms in the context
of bagged features (i.e. $m'$ features are randomly drawn instead of all
$m$ features, and bagging (bagged records).
Sliq/D contains a complex expensive implementation-dependent coordination between workers (in particular for numerical features) which is not detailed here.
$M$ is the maximum number of nodes per depth.
We assume classification. $C$ is the number of nodes in the tree.
$PS$ refers to the complexity of presorting features, see  Lemma \ref{externalsort}.
$[x]$ refers to the size of storage for an $x$ (e.g. $[int]$ refers to the number of bits per integer and $[value]$ refers
to the number of bits for storing one entry of one feature or one label).
$Z$ is defined in Section \ref{maths}, and depends on $z$, $w$, $m$ and $d$; it should be averaged over the depth levels. $z=1$ means that
we decide to use the same $m'$ random features for all nodes of a layer (USB), otherwise $z$ is the maximum number of
nodes in a given depth level.
See Eq. \ref{kp} for more on $Z$.
If conditions of Lemma \ref{hypergeom} are met, then $Z=O( \min(K, \lceil zm'/w\rceil))$.
$K$ is $\lceil m/w\rceil$.
}
}{\caption{\label{comptable}Complexities of some discussed algorithms in the context
of bagged features (i.e. $m'$ features are randomly drawn instead of all
$m$ features, with $m'$ typically scaling as $\sqrt{m}$) and bagging (bagged records).
Sliq/D contains a complex expensive implementation-dependent coordination between workers (in particular for numerical features) which is not detailed here.
$M$ is the maximum number of nodes per depth.
We assume classification; $C$ is the number of nodes in the tree.
$PS$ refers to the complexity of presorting features.
$[x]$ refers to the size of storage for an $x$ (e.g. $[int]$ refers to the number of bits per integer and $[value]$ refers
to the number of bits for storing one entry of one feature or label).
$Z$ is defined in Section \ref{maths}, and depends on $z$, $w$, $m$ and $d$; it should be averaged over the depth levels.
If conditions of Section \ref{maths} are met, then $Z=O(\lceil \min(K, zm'/w)\rceil)$.
$K$ is $\lceil m/w\rceil$.
}
}\end{table*}
}

\section{Experiments: Artificial Datasets}
\label{sec:experimentation}
\begin{figure*}[t!] \scriptsize
\center
\ifthenelse{\longversion=1}{
\includegraphics[width=.48\textwidth]{\detokenize{artifxpfigs/rf_dim17_1treesnouv.png}}
\includegraphics[width=.48\textwidth]{\detokenize{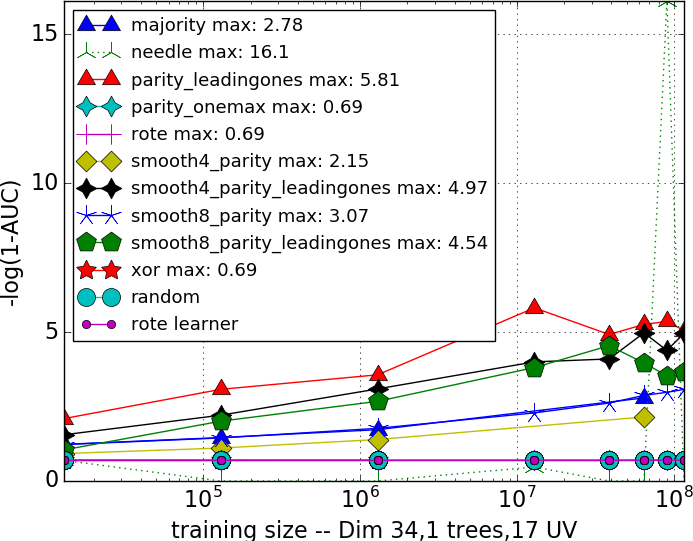}}\\
\includegraphics[width=.48\textwidth]{\detokenize{artifxpfigs/rf_dim17_3treesnouv.png}}
\includegraphics[width=.48\textwidth]{\detokenize{artifxpfigs/rf_dim17_3trees.png}}\\
\includegraphics[width=.48\textwidth]{\detokenize{artifxpfigs/rf_dim17_10treesnouv.png}}
\includegraphics[width=.48\textwidth]{\detokenize{artifxpfigs/rf_dim17_10trees.png}}
}{
\includegraphics[width=.35\textwidth]{\detokenize{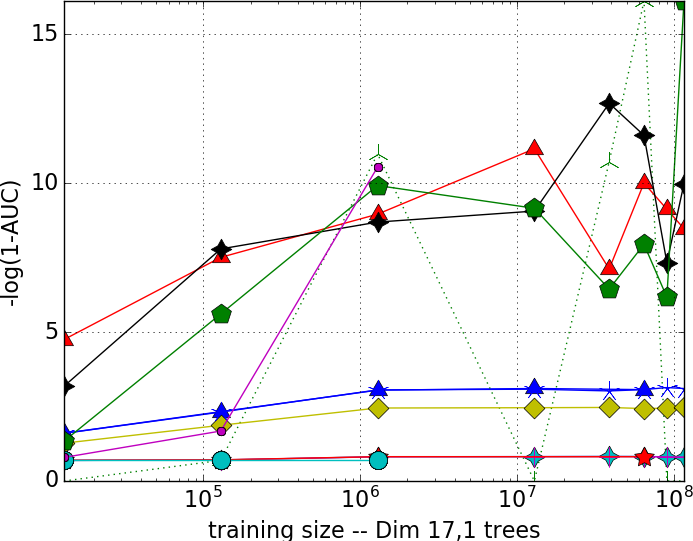}}
\includegraphics[width=.35\textwidth]{\detokenize{artifxpfigs/rf_dim17_1trees.png}}\\
\includegraphics[width=.35\textwidth]{\detokenize{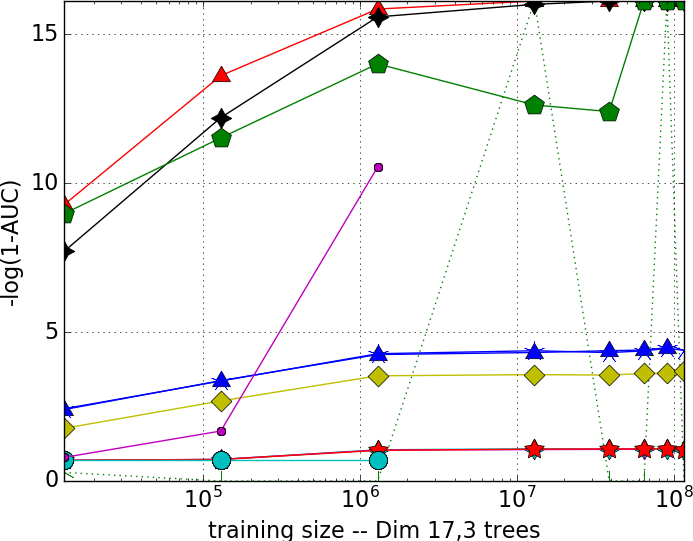}}
\includegraphics[width=.35\textwidth]{\detokenize{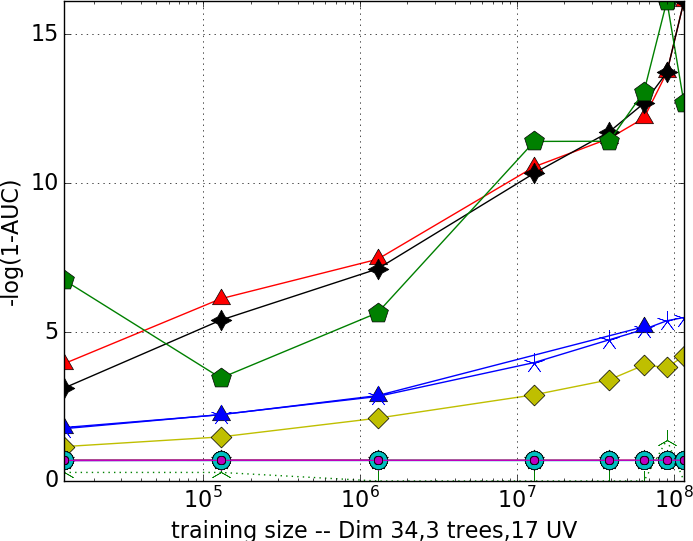}}\\
\includegraphics[width=.35\textwidth]{\detokenize{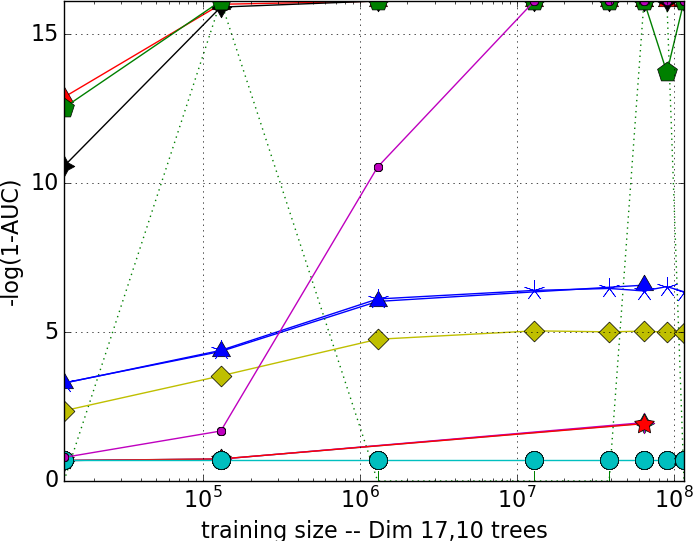}}
\includegraphics[width=.35\textwidth]{\detokenize{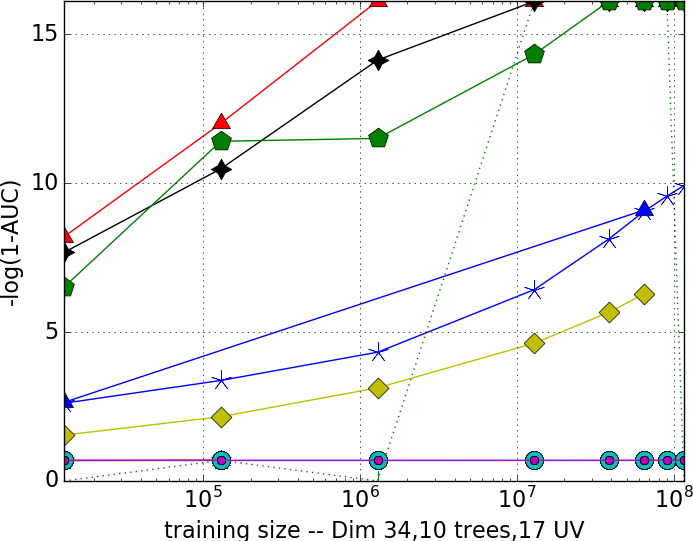}}
}
	\caption{\label{trainingperf2}Impact of the number of trees and training set sizes. Area Under ROC Curve (AUC) on artificial datasets.
	DRF, $m'=\lceil\sqrt{m}\rceil$ randomly drawn features per node, unlimited depth, at least one record per node.
	Similarly to real world data in Section \ref{sec:experimentation2}, even for gigantic datasets, increasing the training set size and/or adding trees helps - in particular with many UV (compare rows, for each column). Rote learning fails (AUC=$\frac12$) when we have UV.
	Random labelling, or labelling according to majority class, leads to $AUC=\frac12$, hence $-log(1-AUC)=log(2)$. One independent run per point in the plot; hence the highly imbalanced ``needle'' (dashed line) leads to irregular curves, others are more stable.
  }
\end{figure*}

\begin{figure*}[t!] \scriptsize
\center
\ifthenelse{\longversion=1}{
\includegraphics[width=.48\textwidth]{\detokenize{artifxpfigs/rf_yaxistime_dim17_3treesnouv.png}}
\includegraphics[width=.48\textwidth]{\detokenize{artifxpfigs/rf_yaxistime_dim18_1treesnouv.png}}\\
\includegraphics[width=.48\textwidth]{\detokenize{artifxpfigs/rf_yaxistime_dim21_1treesnouv.png}}
\includegraphics[width=.48\textwidth]{\detokenize{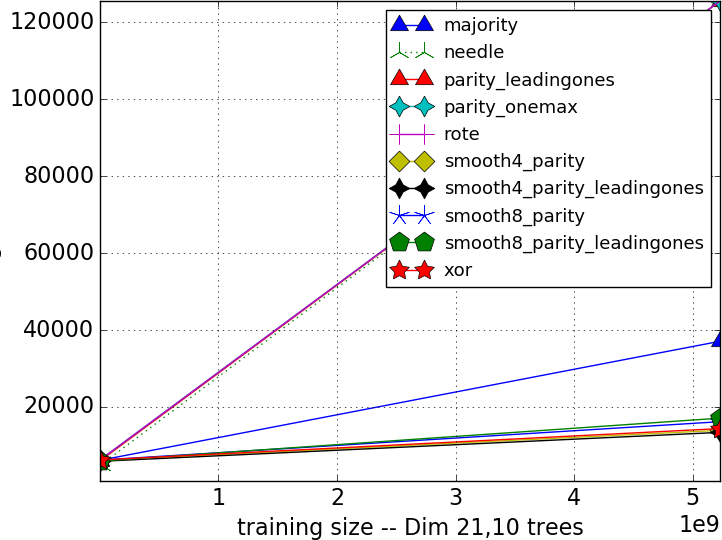}}
}{
\includegraphics[width=.40\textwidth]{\detokenize{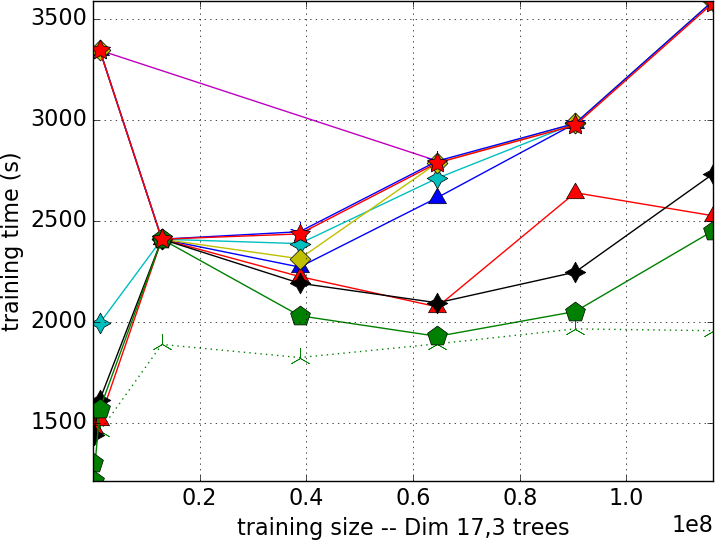}}
\includegraphics[width=.40\textwidth]{\detokenize{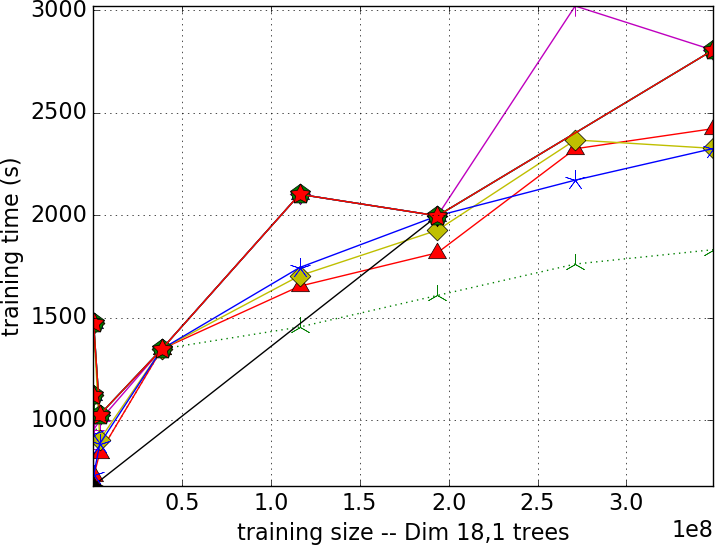}}\\
\includegraphics[width=.40\textwidth]{\detokenize{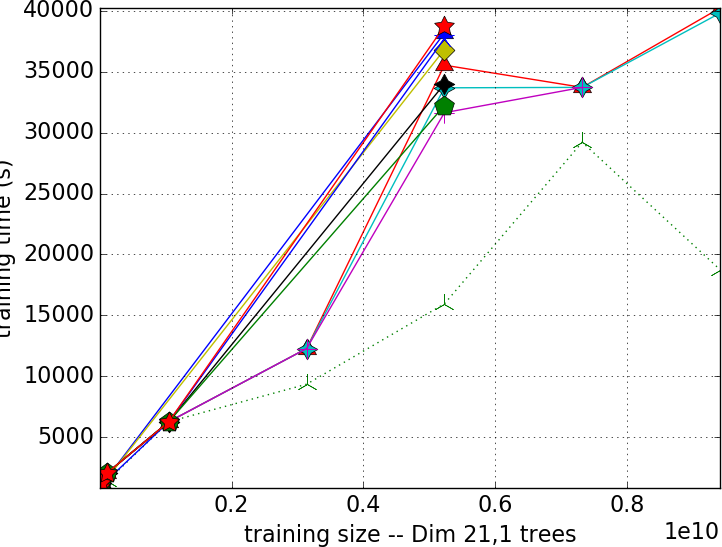}}
\includegraphics[width=.40\textwidth]{\detokenize{artifxpfigs/rf_yaxistime_dim21_10treesnouv.png}}
}
	\caption{\label{trainingtime}Training time in seconds as a function of training set size.
	Exact random forest, $m'=\lceil\sqrt{m}\rceil$ randomly drawn features per node, unlimited depth, at least one record per node.
  We have e.g. 1900s - 3000s for building one random forest tree on 3e8 examples in dimension 18.
  This is in an environment with preemptions; hence irregular results. The number of workers is equal to the dimension,
  independently of the number of trees - the different trees are built sequentially, only the presorting is amortized.
  }
\end{figure*}

In this section, we report the performance of DRF on a set of families of synthetic binary
classification datasets published specifically for large scale machine learning\cite{syntheticdatasets}.
Each family is associated with a ground truth function (e.g. XOR, Majority).
The members of each family differ in the number of training samples as well as the number of informative and
uninformative features. The datasets include various numbers of useless variables (UV), with no correlation with the labels.
Rote learning is used as a baseline for comparison; it consists in just labelling a test sample correctly if it was
in the training set, and randomly otherwise.
We test DRF with hyperparameters as follows: 1, 3 or 10 trees; unbounded depth; minimum number of examples per leaf equal to $1$;
number of splitters equal to the number of features. Given the large number of datasets, we did not replicate any of the experiments;
but each data point is obtained independently.
\ifthenelse{\longversion=1}{

}{}
These runs are performed with a low priority - this shows that the approach remains reliable in spite of
interruptions (workers can be killed by tasks with higher priority).
\ifthenelse{\longversion=1}{
Figures \ref{trainingperf1} and \ref{trainingperf2}
}{
Figure \ref{trainingperf2} and SM
}
show the AUC as a function of the training set size, while Figure \ref{trainingtime} shows the training time.
\ifthenelse{\longversion=0}{}{
\begin{figure*}[t!] \scriptsize
\includegraphics[width=.48\textwidth]{\detokenize{artifxpfigs/rf_dim14_1treesnouv.png}}
\includegraphics[width=.48\textwidth]{\detokenize{artifxpfigs/rf_dim14_1trees.png}}\\
\includegraphics[width=.48\textwidth]{\detokenize{artifxpfigs/rf_dim14_3treesnouv.png}}
\includegraphics[width=.48\textwidth]{\detokenize{artifxpfigs/rf_dim14_3trees.png}}\\
 \includegraphics[width=.48\textwidth]{\detokenize{artifxpfigs/rf_dim14_10treesnouv.png}}
 \includegraphics[width=.48\textwidth]{\detokenize{artifxpfigs/rf_dim14_10trees.png}}\\
	\caption{\label{trainingperf1}Area Under ROC Curve (AUC) on artificial datasets (test dataset) for various numbers of trees and with/without useless variables. Bigger training set sizes in Figure \ref{trainingperf2}.
	Exact random forest, $m'=\lceil\sqrt{m}\rceil$ features per node, unlimited depth, at least one record per node. This is rather small size - the goal of this experiment
  is to validate the testbeds; we see that with useless variables, rote learning performs very poorly; even without, DRF outperforms rote learning unless the training set size is large.
  Please note that the expected AUC performance of rote learning is independent of the dataset. Increasing the number of trees improves the AUC.}
\end{figure*}}
\ifthenelse{\longversion=1}{
\begin{figure*}[t!]\scriptsize
\includegraphics[width=.48\textwidth]{\detokenize{artifxpfigs/rf_dim17_1treesnouv.png}}
\includegraphics[width=.48\textwidth]{\detokenize{artifxpfigs/grf_dim17_1treesnouv.png}}\\
	\caption{\label{ubs} Comparison between DRF and DRF with USB.
  The number $w$ of workers is the number $m$ of features,
  so there is no big computation time improvement to expect - USB aims at reaching the same performance
  with $w\simeq m''$ and $m$ large (Section \ref{maths}) - we just check here the quality of obtained models.
  The main conclusion from these
  results is that USB does not harm AUC on average (we have better performance
  for nearly half results, as visible in the legend),
  though it increases variance.
  The impact of USB in the case $w=m'' << m$ will be investigated experimentally in a further work.}
\end{figure*}}{}
\ifthenelse{\longversion=1}
{
\def\mysw{0.47\textwidth}
\def\myw{1.00\textwidth}
}
{
\def\mysw{0.40\textwidth}
\def\myw{1.00\textwidth}
}
\begin{figure*}[t!]
    \centering
    \begin{subfigure}[t]{\mysw}
        \centering
        \includegraphics[width=\myw]{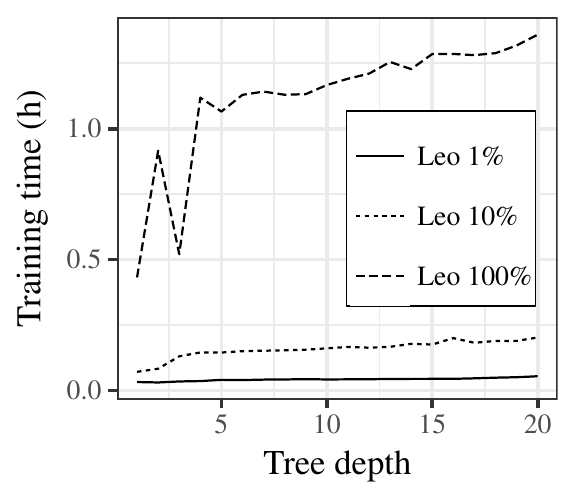}
    \end{subfigure}%
	~
	 \begin{subfigure}[t]{\mysw}
        \centering
        \includegraphics[width=\myw]{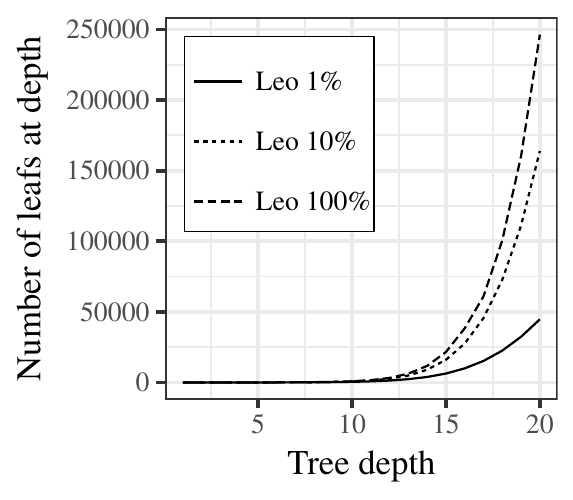}
    \end{subfigure}
  ~
    \begin{subfigure}[t]{\mysw}
        \centering
        \includegraphics[width=\myw]{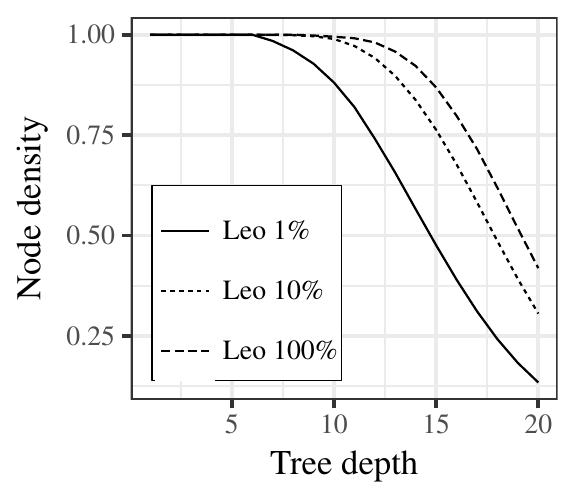}
    \end{subfigure}%
  ~
    \begin{subfigure}[t]{\mysw}
        \centering
        \includegraphics[width=\myw]{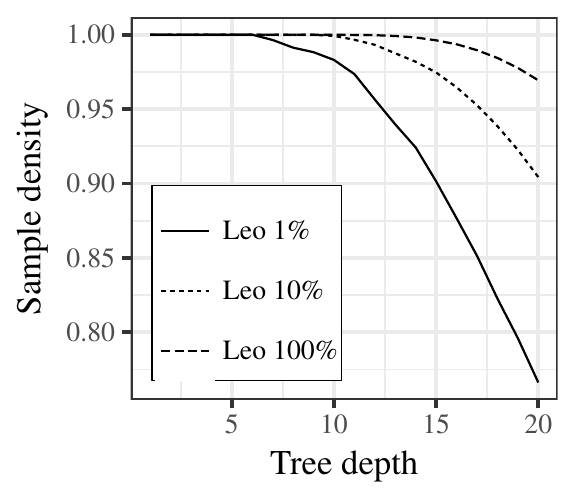}
    \end{subfigure}
    ~
    \begin{subfigure}[t]{\mysw}
        \centering
        \includegraphics[width=\myw]{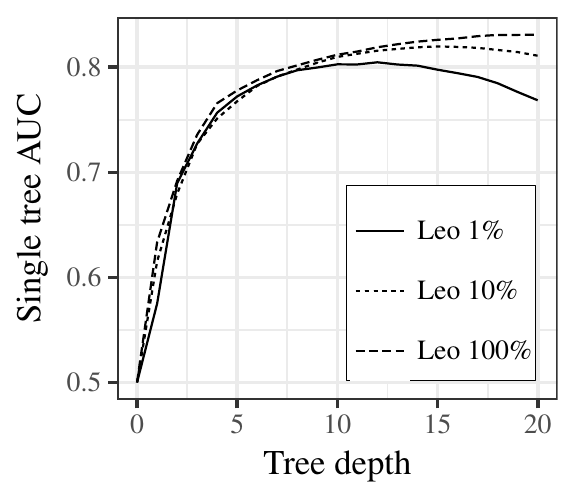}
    \end{subfigure}%
    ~
    \begin{subfigure}[t]{\mysw}
        \centering
        \includegraphics[width=\myw]{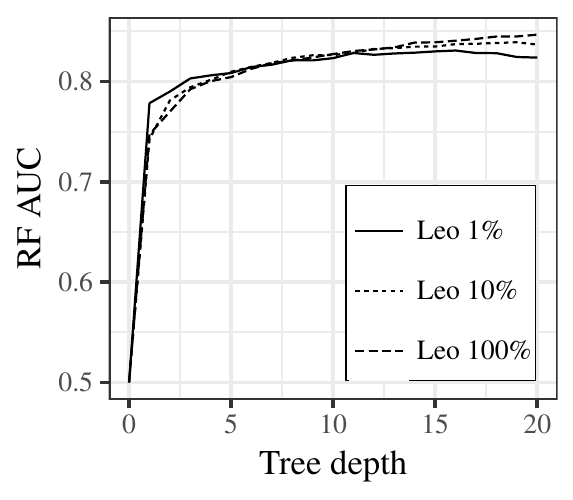}
    \end{subfigure}%
    \caption{Tree and RF metrics while depth-by-depth training. Shows the training time, number of open leaves, node density,  sample density, individual trees' AUC and TF AUC for depth level between 0 and 20.}
	\label{fig:gmob_tree_stats}
\end{figure*}
\def\removedforthemoment{
\subsection{Synthetic Datasets with Numerical Features}
\subsubsection{Julia dataset} is ...
Categorical features with a limited number of possible values can be tackled
much faster by computing histograms (see e.g. \cite{rainforest}). In the present
section, we consider a purely numerical dataset with two classes, namely the Julia
set. We use the following version: $x$ and $y$ are randomly draw uniformly and independently in $(0,1)$.
$z_0(x,y)$ is defined as $2x-1+i*(2y-1)$. Then iterates are defined by $z_{n+1}(x,y) = z_n(x,y)^2+c$,
for $c=0.3+0.5i$. The class $v$ of $(x,y)$ is $1$ if and only if $\max_{i<200} |z_i|\leq 2$.

As well as for previous datasets, we define versions with useless variables having no impact on the label.

TODO results
}

\section{Experiments: Real World Dataset}\label{sec:experimentation2}
\ifthenelse{\longversion=1}{
In this section we report and discuss the performance of DRF on a set of large datasets.
\subsection{Leo dataset}
}{}
The \emph{Leo dataset} is a large unbalanced proprietary binary classification dataset containing 18 billions records. Each record is defined by 3 numerical features and 69 categorical features with respective arities ranging from 2 to 10'000.
To put the size of this dataset into perspective, storing a single 8 byte integer for each of the samples (e.g. the index of a sample) would requires 114 gigabytes of memory. By comparison, high end consumer computer stations have between 8 and 16 gigabytes of memory. If densely represented and uncompressed (and assuming zero overhead for the dataset structure), the dataset occupies 6 terabytes of memory. As far as we know, the Leo dataset is the largest dataset ever used to train an (exact or approximate) Random Forest.
We consider three versions of this dataset: Leo 1\% and Leo 10\%, and Leo 100\%, respectively 1\%, 10\%, and 98\% of the full dataset.
We reached an UAC of .847. The best AUC on this dataset (obtained by deep learning) was 0.81.
\ifthenelse{\longversion=1}{
\subsection{Experimental Setup}}{}
We apply the DRF algorithm on the datasets described above.
The number of workers is set to 82.
Small and moderate size subsets of datasets can be run with the dataset loaded entirely in memory (distributed across the workers). Reading datasets from memory is significantly faster than reading from drive. However, for the sake of the comparison, all experiments have been run with the datasets remaining on drive.
In all the runs, the hyperparameters of the Random Forest have been set to some reasonable default values: the number of candidate attributes for each split is equal to the square root of the total number of attributes, the minimum number of records in each node is set to 10, 100 and 1000 respectively, and the maximum depth is set to 20. In the case of subsets of dataset, the minimum number of records has been reduced proportionally with the relative size of the subset to the original set.
\ifthenelse{\longversion=1}{
\subsection{Results}
}{}
Table~\ref{tab:times} shows training time, number of nodes, node density and sample density of each tree (averaged over all the trees in the model).
The \emph{node density} is the ratio between the number of leaves of the tree and the number of leaf of a dense tree of similar
depth (i.e. $2^D$ with $D$ the depth of the tree). The \emph{sample density} is the ratio of training samples that reached the bottom leaves of the tree (i.e. the leaves at depth 20). Both density measures are expected to decrease with the depth of the tree.
\begin{table}
\centering
\caption{Average training time, number of nodes, node density and sample density for the various datasets.}
\label{tab:times}
\label{my-label}
\tabcolsep=0.11cm
\scriptsize
\begin{tabular}{|p{1.0cm}|p{1.3cm}|p{1.0cm}|p{1.2cm}|p{1.0cm}|p{1.1cm}|}
\toprule
Leo & Samples & Train time (h) & Leaves & Node density & Sample density \\
\midrule
 1\% & $.173\cdot 10^9$ & 0.838 & $140\cdot 10^3$ & 0.134 & 0.766 \\
 10\% & $1.73\cdot 10^9$ & 3.156 & $320\cdot 10^3$ & 0.305 & 0.904 \\
 100\% & $17.3\cdot 10^9$ & 22.29 & $435\cdot 10^3$ & 0.415 & 0.969 \\
\bottomrule
\end{tabular}
\end{table}
Figure~\ref{fig:gmob_tree_stats} shows the average training time, number of leaves, node density and sample density, when varying the maximum tree depth between 0 (i.e. a tree is a single root node attached to the majority class) and 20.
Figures have been averaged across the trees of a RF.
The total training time of a tree is the sum of the training times of each depth levels.
As expected, the number of leaves and the training time increases with the depth of the trees as the number of leaves grows.
However, while the number of leaves increases exponentially with the depth of the trees, the computation time does not. This is explained by the fact that most of computation is spent on scanning the dataset, and this step does not depend on the number of leaves.
A depth 20, in the case of Leo 100\%, 96.9 \% of the training samples are still in
an open leaf (i.e. a leaf that could be split if the maximum depth was increased).
This indicates that the tree can still grow if the maximum depth is increased. This also shows that pruning the dataset by removing samples in closed leaves
	\ifthenelse{\longversion=1}{
		(as in Sprint; see also~\ref{sec:best_of_two_words})}{(as in Sprint)}
	would not speed up the computation since the cost of pruning would exceed its gains (using a pruning as in Sprint would not provide any significant improvement given this 96.9\%).
Figure~\ref{fig:gmob_tree_stats} also shows the AUC (Area Under the Receiver Operating Characteristic Curve;
computed on a test set) of individual trees (averaged over several trees) and of the entire RF model when varying the maximum tree depth.
We see that using several billions of examples is useful for improving the AUC.
Non-pruned DTs are highly susceptible to overfitting. Among other causes, DT overfitting appears when the number of training samples in a node becomes too small. Then, the tree starts ``learning the noise'' of the training dataset, the test AUC of an individual tree decreases, and the AUC of the overall RF plateaus.
The depth of a tree is the main factor leading to nodes with few training samples.
Therefore, we expect for the effect described above to be correlated with the depth of the trees:
in the case of Leo 1\% and Leo 10\%, the overfitting of individual trees starts respectively at depth 13 and 17;
in the case of Leo 100\%, the overfitting of individual trees has not yet started at depth 20.
The overfitting of individual trees does not indicate that the overall RF is not expected to benefit from deeper training: the AUC of the corresponding RFs plateaus at depth 16 for Leo 1\%, and keeps increasing after depth 20 for Leo 10\% and Leo 100\%.
We also observe that the RF trained on more data are plateauing to greater AUCs (0.823, 0.837, and 0.847, respectively for Leo 1\%, Leo 10\%, and Leo 100\%).
This indicates that RF benefits from using large datasets and training deeper trees.



\section{Conclusions}
\label{sec:conclusion}
We introduced DRF, an exact distributed Random Forest~\cite{randomforest}.
Our method stands out from existing exact distributed approaches by a smaller space, disk and network complexity. We demonstrate its application to 17.3B training examples with dense features. This figure is 1000x larger than any exact decision tree from RF or gradient-boosted trees found in the literature~\cite{tonghexgboost,distxgboost,cudt}, and at least 10x larger than any related approximate work~\cite{dividedata,boat,rfbigdata,bigrf1}.
It is known that more training data improves accuracy or AUC;
however, it is not obvious that this is the case
when increasing the number of training examples from 1 to 10 billions.
This is also central to learning algorithms working on subsets of the dataset~\cite{boat}.
The present results show examples of
datasets (both artificial and real world) for which more training data is beneficial
in case of datasets of billions of examples.
\ifthenelse{\longversion=1}{

\subsection*{Further work}
 The analysis has shown two elements which were not exploited in the present experimental results:
 \begin{itemize}
	 \item Redundancies over features (i.e. store each feature on two workers and
   dynamically choose the worker with the least load for working on it) can reduce
   the overall complexity
		 \ifthenelse{\longversion=1}{(corollary \ref{coreg}); }{(Section \ref{maths})}
		 this was not experimented in the present work
   which considers large numbers of records rather than large numbers of features.
   This will not improve the complexity if $m=w$, but for large numbers of features $m=w$ might be
   impossible and using redundancies we might get an almost constant complexity with $w$ equal
   to the number of features which are considered in a given layer (i.e. $\min(zm',m)$ in the terminology
   of Table \ref{comptable}).
	 \item The factor $Z$ (maximum number of features of a worker for a given depth), and therefore the USB modification used in some RF implementations,
         has a strong impact on the number of machines needed for achieving a logarithmic cost as a function
         of the number of features (essentially switching from $w=m$ to $w=m'$, in a RF context where usually $m'\simeq \sqrt{m}$). The
            present paper is dedicated to a strictly equivalent random forest, hence this is left as further work.
\end{itemize}
We can also improve the implementation by switching to pure memory for nodes with a moderate number of records.}{
{\bf{Further work.}} The mathematical analysis has shown the importance of USB and of redundant storage of features, to be
experimentally investigated; we might also switch to pure memory for nodes with small numbers of records.
}

\ifthenelse{\longversion=1}
{
		\bibliographystyle{abbrv}

	}{
		\bibliographystyle{icml2018}}
\bibliography{references}

\end{document}